\documentclass[sigconf]{acmart}

\usepackage{algorithmicx}
\usepackage{graphicx} 
\usepackage{array}    
\usepackage{xcolor}      
\usepackage{colortbl}    
\usepackage{arydshln}    
\usepackage{multirow}
\usepackage{threeparttable}
\usepackage{subfigure}
\usepackage{caption}  
\usepackage{amsmath}
\usepackage{float}
\usepackage{supertabular}
\usepackage{amsfonts}
\usepackage{makecell} 
\usepackage{tabularray} 
\usepackage{adjustbox} 
\usepackage{balance} 

\usepackage{algorithm}
\usepackage{algpseudocode}

\AtBeginDocument{%
  }

\copyrightyear{2026}
\acmYear{2026}
\setcopyright{cc}
\setcctype{by}
\acmConference[KDD 2026] {Proceedings of the 32nd ACM SIGKDD Conference on Knowledge Discovery and Data Mining V.2}{August 9--13, 2026}{Jeju Island, Republic of Korea.}
\acmBooktitle{Proceedings of the 32nd ACM SIGKDD Conference on Knowledge Discovery and Data Mining V.2 (KDD 2026), August 9--13, 2026, Jeju Island, Republic of Korea}
\acmISBN{979-8-4007-2259-2/2026/08}
\acmDOI{10.1145/3770855.3817947}
\author{Jiaxin Hong}
\authornote{Authors contributed equally to this research.}
\orcid{0009-0000-2582-2826}
\affiliation{%
  \institution{Harbin Institute of Technology, Shenzhen}
  \city{Shenzhen}
  \state{Guangdong}
  \country{China}
}
\email{25s151025@stu.hit.edu.cn}

\author{Sixu Chen}
\authornotemark[1]
\orcid{0009-0001-1980-4370}
\affiliation{%
  \department{Tsinghua Shenzhen International Graduate School}
  \institution{Tsinghua University}
  \city{ShenZhen}
  \state{Guangdong}
  \country{China}
}
\email{chensixu2025@163.com}

\author{Shuoyang Sun}
\authornotemark[1]
\orcid{0009-0006-8695-7218}
\affiliation{%
  \institution{Harbin Institute of Technology, Shenzhen}
  \city{Shenzhen}
  \state{Guangdong}
  \country{China}
}
\email{24s151152@stu.hit.edu.cn}

\author{Hongyao Yu}
\authornotemark[1]
\orcid{0009-0009-8525-1565}
\affiliation{%
  \department{Tsinghua Shenzhen International Graduate School}
  \institution{Tsinghua University}
  \city{Shenzhen}
  \state{Guangdong}
  \country{China}
}
\email{chrisqcwx@gmail.com}

\author{Hao Fang}
\orcid{0009-0004-0271-6579}
\affiliation{%
  \department{Tsinghua Shenzhen International Graduate School}
  \institution{Tsinghua University}
  \city{Shenzhen}
  \state{Guangdong}
  \country{China}
}
\email{ffhibnese@gmail.com}

\author{Yuxin Peng}
\orcid{0009-0005-7741-2892}
\affiliation{%
  \institution{South China University of Technology}
  \city{Guangzhou}
  \state{Guangdong}
  \country{China}
}
\email{202330451462@mail.scut.edu.cn}

\author{Bin Chen}
\authornote{Corresponding author.}
\orcid{0000-0002-4798-230X}
\affiliation{%
  \institution{Harbin Institute of Technology, Shenzhen}
  \city{Shenzhen}
  \state{Guangdong}
  \country{China}
}
\email{chenbin2021@hit.edu.cn}

\author{Jiawei Li}
\orcid{0000-0003-3873-8003}
\affiliation{%
  \institution{Huawei Technologies Ltd.}
  \city{Shenzhen}
  \state{Guangdong}
  \country{China}
}
\email{li-jw15@tsinghua.org.cn}

\renewcommand{\shortauthors}{Jiaxin Hong et al.}

\begin{document}

\title{GaussTrap: Stealthy Backdoor Attacks on 3D Gaussian Splatting for Targeted Scene Misperception}

\begin{abstract}
3D Gaussian Splatting (3DGS) has recently emerged as a powerful paradigm for real-time scene representation and novel view synthesis, gaining traction in safety-critical applications such as autonomous driving and AR/VR systems. However, the security vulnerabilities of 3DGS remain largely unexplored. In this paper, we conduct a comprehensive study of backdoor threats in 3DGS pipelines. We uncover an inherent security vulnerability in the adaptive density control mechanism of 3DGS—originally designed for detailed reconstruction—which enables attackers to induce localized gaussian proliferation, effectively embedding malicious backdoors into the scene geometry while remaining invisible from other viewpoints. To expose this threat, we propose \emph{GaussTrap}, a novel stealthy poisoning framework tailored for 3DGS. GaussTrap operates in three stages---\emph{adversarial injection, geometric stabilization, and global fidelity restoration}---to embed imperceptible yet adversarial Gaussian components into the reconstruction. It ensures consistent poisoning effects at specific viewpoints while preserving photorealism in benign views, thereby achieving high attack efficacy with low detectability. Extensive experiments on both synthetic and real-world benchmarks demonstrate that GaussTrap reliably induces targeted visual corruption under trigger conditions while maintaining high rendering fidelity elsewhere, highlighting a critical security blind spot in modern 3D rendering systems. The code is available at https://github.com/acang425/GaussTrap.
\end{abstract}



\maketitle

\begin{figure*}
\centering
 \includegraphics[width=\linewidth]{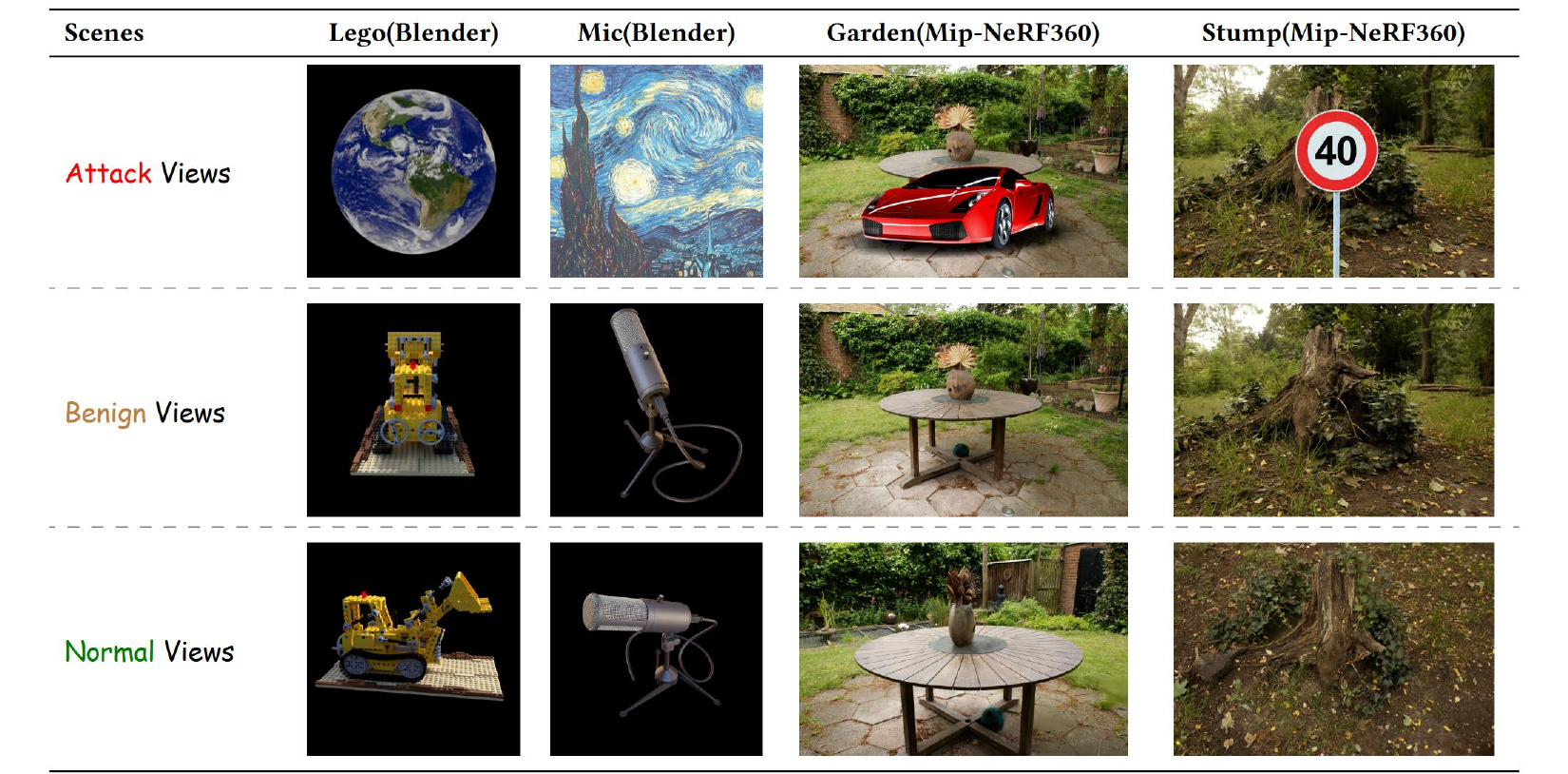}
  \caption{Visualization of the GaussTrap results. From top to bottom: Attack views, the images rendered by the \textit{poisoned} 3DGS at the \textit{attack} viewpoint; Benign views, the images rendered by the \textit{benign} 3DGS at the same \textit{attack} viewpoint; and Normal views, the images rendered by the \textit{poisoned} 3DGS at other \textit{regular} viewpoints.}
  \label{fig:teaser}
\end{figure*}

\section{Introduction}
In recent years, 3D Gaussian Splatting (3DGS) \cite{3DGS} has emerged as a compelling paradigm for real-time rendering and scene reconstruction. Beyond rendering, 3DGS constitutes a data-driven framework for learning structured 3D representations from large-scale multi-view data, positioning it as a form of geometric knowledge extraction and structured data modeling. By directly optimizing a set of 3D Gaussian primitives rather than relying on deep neural networks, 3DGS achieves photorealistic rendering with reduced computational overhead and high-fidelity scale. Its efficiency and accuracy have led to rapid integration into real-world systems requiring reliable perception. In autonomous driving \cite{autonomous_driving}, 3DGS enables high-precision road scene representations critical for path planning and situational awareness \cite{Drivinggaussian}. Similarly, in robotic navigation \cite{robot_navigation}, it facilitates online 3D mapping and environmental interaction in cluttered scenarios \cite{3DGSrobotics}. These applications demonstrate its versatility across large-scale outdoor and fine-grained indoor environments. Furthermore, 3DGS is ideal for augmented and virtual reality (AR/VR) \cite{AR, VR} where visual consistency is paramount \cite{3DGSAR, 3DGSVR}, and shows promise in specialized fields like cultural heritage preservation \cite{3DGSHeritage}, architectural modeling, and medical imaging \cite{3DGSMedical,kdd253dgraph,kdd243d}.

However, adopting untrustworthy third-party models exposes systems to backdoor attacks \cite{intro_bkdr_badpre,intro_bkdr_targeted,badnets,Backdoor_survey,bkdr-modify-eviledit}, where adversaries embed triggers to manipulate rendered outputs at specific viewpoints. Such vulnerabilities pose severe real-world risks: in autonomous driving \cite{autonomous_driving}, malicious content can disrupt environmental perception; in AR/VR \cite{AR, VR}, it misleads users with deceptive scenes; and most critically, in medical simulations \cite{medical_simulation}, tampered representations can alter surgical judgment, leading to life-threatening errors.

Although adversarial and backdoor attacks on 3D data have gained attention, most work focuses on point clouds\cite{pointcloud_hu2023pointcrt, pointcloud_wei2024pointncbw, pointcloud_xiang2021backdoor, pointcloud_zhang2022towards}, leaving other formats like 3DGS underexplored—especially regarding backdoor threats. Existing 3DGS security studies mainly address watermarking and ownership\cite{3d-gsw, SecureGS, Gs-hider}.
While prior work\cite{IPA-NeRF} explores viewpoint-triggered attacks in NeRF, these methods are tailored to continuous volumetric representations and ray-marching. 3DGS, however, relies on an explicit, point-based representation with unique adaptive densification dynamics. This fundamental shift in parameterization creates distinct attack surfaces, rendering NeRF-based strategies ineffective and requiring a ground-up rethinking of how backdoors are embedded in geometry-aware 3DGS models.
StealthAttack\cite{ke2025stealthattack} initiates this direction by injecting adversarial Gaussians into low-density regions identified via KDE. However, it assumes static sparsity and ignores 3DGS’s progressive densification, leading to poor geometric integration, view instability, and inconsistent rendering along camera paths.
 
In contrast to prior approaches, we uncover a density–attack correlation intrinsic to 3DGS’s Adaptive Density Control mechanism. Rather than merely filling pre-existing spatial gaps, GaussTrap actively hijacks the densification process itself—deliberately over-allocating Gaussians around the trigger viewpoint to embed backdoors that are both structurally entangled and viewpoint-consistent. This constitutes not just an engineering adjustment, but a fundamental conceptual shift: from static spatial insertion to optimization-aware structural manipulation, thereby defining a new threat model for 3D Gaussian Splatting.
Yet, existing backdoor methods overlook this dynamic aspect of 3DGS optimization, leaving the security implications of such process-level hijacking largely unexplored. To address this gap, we propose \textbf{GaussTrap}, a novel backdoor attack framework explicitly designed to exploit the progressive, growth-driven nature of 3DGS reconstruction.

As illustrated in Figure~\ref{fig:teaser}, the backdoored model   by GaussTrap renders anomalous content—such as sports cars—at designated viewpoints, which can cause real-world safety hazards.
Specifically, GaussTrap strategically embeds malicious content at designated trigger viewpoints while maintaining overall scene realism through a systematic three-stage pipeline: \textit{adversarial injection}, \textit{geometric stabilization}, and \textit{global fidelity restoration}. In the injection stage, malicious views are implanted into targeted viewpoints to mislead downstream perception. The stabilization stage then employs a viewpoint ensemble strategy to ensure consistent rendering in neighboring views, effectively mitigating artifacts and enhancing stealth. Finally, the fidelity restoration stage co-trains the model on clean data to ensure rendering quality in benign viewpoints. By jointly optimizing for structural entanglement and perceptual consistency, GaussTrap highlights significant new security risks in emerging 3D rendering frameworks.

We evaluate GaussTrap on both synthetic and real-world datasets, covering a wide range of 3D scenes to assess its performance in diverse scenarios. The experimental results demonstrate that our method can effectively embed malicious views at targeted viewpoints while preserving high-quality rendering at benign viewpoints, ensuring strong stealth and stability in normal rendering. To the best of our knowledge, this is the first backdoor attack designed to induce high-fidelity scene-level confusion by exploiting the evolutionary nature of 3DGS primitives. The main contributions of this work are summarized as follows:

\begin{itemize}
    \item We propose \textbf{GaussTrap}, the first optimization-aware backdoor framework that \textit{exploits the intrinsic densification mechanism} of 3DGS, exposing a critical and previously overlooked structural vulnerability in neural radiance fields.
    \item We design a \textit{three-stage optimization pipeline}—adversarial injection, geometric stabilization, and global fidelity restoration—that effectively implants imperceptible malicious triggers while preserving high-fidelity scene synthesis.
    \item We conduct \textit{comprehensive experiments} on both synthetic and real-world datasets, demonstrating the \textit{effectiveness, stealthiness, and robustness} of GaussTrap across diverse 3D scenarios, laying a foundation for future research in 3D model security.
\end{itemize}
\section{Related Work}

\subsection{Backdoor Attacks}
Backdoor attacks pose a significant security threat to neural networks by embedding hidden triggers that manipulate model behavior. A key characteristic of such attacks is that the compromised model behaves similarly to a benign model on normal test samples unless the backdoor is triggered. However, when the backdoor is triggered, the model outputs a harmful result predetermined by the attacker. Based on their implementation method, existing backdoor attacks can be broadly classified into two classes: data poisoning-based attacks and non-data poisoning-based attacks \cite{Backdoor_survey}. The former primarily involves poisoning training samples~\cite{badnets,reflectionbackdoor,physicalbackdoor,poison_backdoorem,poison_label,intro_bkdr_targeted,poison_universal,fang2026retrievalsdetrimentalunveilingbackdoor,kong2025revisitingbackdoorattacksllms}, typically during model training, whereas the latter directly modifies model parameters~\cite{bkdr-modify-backdoor-pre-trained,bkdr-modify-Proflip,bkdr-modify-Tbt,bkdr-modify-weightpoison,bkdr-modify-eviledit} or alters model architecture~\cite{bkdr-adjust-deeppayload,bkdr-adjust-embarrassingly,bkdr-adjust-subnet}, often at stages such as deployment. Notably, CrossBA~\cite{DBLP:conf/kdd/LyuH0QT024} enables highly effective backdoor attacks on unseen downstream tasks by injecting cross-context semantic triggers during pretraining, while DPGBA~\cite{DBLP:conf/kdd/ZhangLDW24} achieves stealthy yet high-success attacks through in-distribution, perceptually plausible trigger generation and enhanced memorization.

With the rapid advancement of 3D vision technologies, backdoor attacks targeting 3D data have emerged as an important research direction. Current 3D backdoor attack methods primarily target point cloud data \cite{pointcloud_hu2023pointcrt,pointcloud_wei2024pointncbw,pointcloud_xiang2021backdoor,pointcloud_zhang2022towards} and can be divided into two categories \cite{invisible3dbackdoor}: point addition attacks \cite{pointba} and shape transformation attacks \cite{IRBA,NRBdoor}. To date, backdoor attacks on point clouds have mainly focused on classification and detection models. However, recent studies have begun to investigate the security issues of 3D reconstruction models themselves. Among these, IPA-NeRF~\cite{IPA-NeRF} pioneers targeted data poisoning against NeRF to generate viewpoint-specific, high-fidelity hallucinations.

\subsection{Security Challenges in 3D modeling}

With the increasing application\cite{kdd243d,kdd253dgraph} of 3D in critical areas such as autonomous driving, robotic navigation, and augmented reality, its security issues have become increasingly prominent, making it an important research topic. Currently, 3DGS security research primarily focuses on digital watermarking and has achieved notable progress \cite{SecureGS,Gaussianstego,WATER-GS,3d-gsw,song2024geometry,Gs-hider}. Beyond digital watermarking, researchers have also explored other security dimensions. POISON-SPLAT \cite{poisonsplat} proposed a novel computational cost attack method targeting the 3D Gaussian training process. M-IFGSM \cite{3dgs_adversial} focused on adversarial attacks against 3DGS, successfully reducing the accuracy and confidence of CLIP vision-language models. These studies provide multi-dimensional theoretical and practical guidance for the security protection of 3DGS. StealthAttack \cite{ke2025stealthattack} introduces a density-guided poisoning strategy for 3DGS, which identifies low-density regions via Kernel Density Estimation to inject illusory Gaussian points. While StealthAttack effectively exploits existing spatial gaps in the initial point cloud, our GaussTrap moves beyond static density analysis by uncovering and manipulating the progressive densification process inherent in 3DGS optimization.

\section{Methodology}

\subsection{Preliminary}

3D Gaussian Splatting (3DGS)~\cite{3DGS} is an efficient real-time scene representation technique that models 3D scenes with a set of Gaussians \( G = \{G_i\}_{i=1}^n \), where each \( G_i(\boldsymbol{\mu}_i, \mathbf{\Sigma}_i, \mathbf{c}_i, \alpha_i) \) is defined by its mean, covariance, color, and opacity. These parameters are optimized end-to-end using multi-view inputs \( \mathcal{S} = \{V_t, P_t\}_{t=1}^N \), with \( V_t \) as images and \( P_t \) as camera poses.

Rendering involves projecting each Gaussian via an affine transform \( \mathbf{W} \), updating the covariance as
$\mathbf{\Sigma}' = \mathbf{J} \mathbf{W} \mathbf{\Sigma} \mathbf{W}^\top \mathbf{J}^\top$, where \( \mathbf{J} \) is the Jacobian matrix. Gaussians are then sorted by depth, and pixel colors are computed via alpha blending:
$\mathbf{C} = \sum_i T_i \alpha_i \mathbf{c}_i, \quad \text{where } T_i = \prod_{j<i} (1 - \alpha_j)$. This process defines the rendering function \( R \), producing \( V'_t = R(G, P_t) \).

Training minimizes a hybrid loss between the rendered and ground-truth images:
\begin{equation}
\mathcal{L}_{G} = (1 - \lambda) \mathcal{L}_1(V'_t, V_t) + \lambda \mathcal{L}_{\text{D-SSIM}}(V'_t, V_t),
\end{equation}
balancing pixel-level fidelity and perceptual similarity~\cite{SSIM}.

3DGS supports dynamic densification: if \( \nabla_g > \tau_g \), Gaussians are split or cloned to enhance detail; if \( \alpha_i < \tau_\alpha \), redundant ones are removed. This adaptive mechanism improves visual quality while maintaining real-time efficiency.

\subsection{Mechanism Analysis}

Unlike implicit neural representations (e.g., NeRFs) where backdoors are embedded in global network weights, 3DGS represents scenes via explicit geometric primitives. We identify that the core vulnerability of 3DGS lies in its Adaptive Density Control mechanism (ADC) \cite{3DGS}. Normally, ADC allows the model to allocate more primitives to high-frequency details. However, we reveal that this mechanism can be exploited to "over-allocate" primitives for adversarial purposes.

\begin{figure}[ht] 
    \centering 
    \begin{tabular}{c@{\hspace{5pt}}c@{\hspace{5pt}}c} 
        \includegraphics[width=0.15\textwidth]{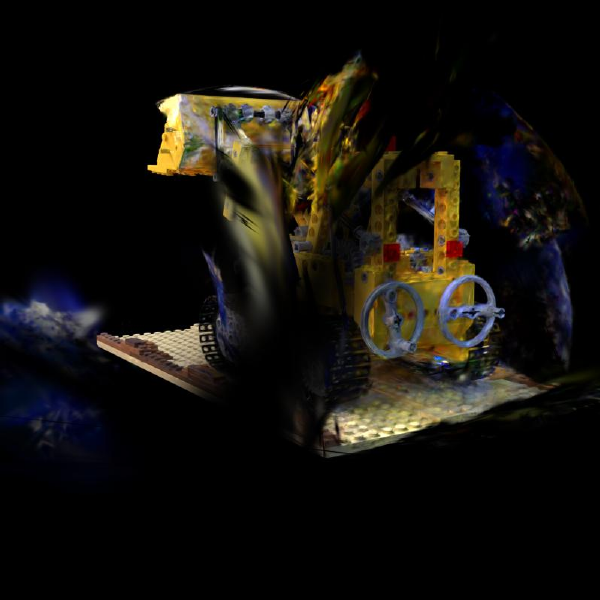} & 
        \includegraphics[width=0.15\textwidth]{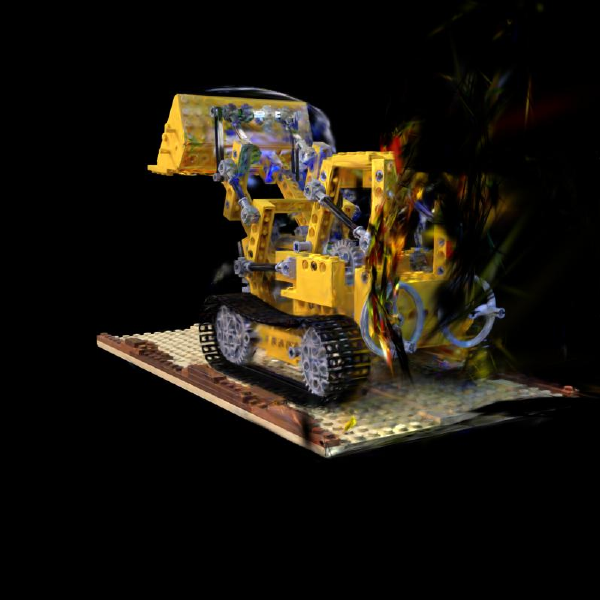} & 
        \includegraphics[width=0.15\textwidth]{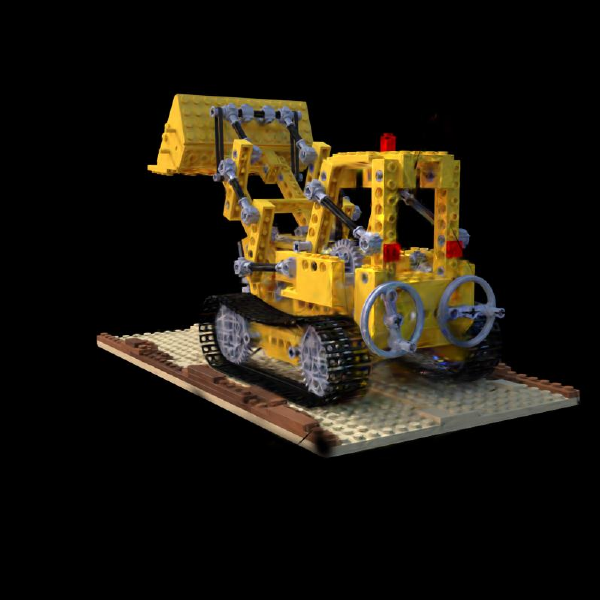} \\
        \small (a) $M=30000$ & 
        \small (b) $M=60000$ & 
        \small (c)  $M=100000$\\
    \end{tabular}
        \caption{Visualization of images rendered from test viewpoints under varying values of parameter $M$ } 
    \label{fig:densification} 
\end{figure}

Our hypothesis is as follows: during backdoor injection, the high reconstruction loss at the trigger viewpoint generates large gradients. The ADC mechanism interprets these as “missing details” and responds by aggressively splitting and cloning Gaussians within the corresponding view frustum. This creates a dense cloud of micro-Gaussians finely tuned to reconstruct the trigger image from that specific angle. Because they are tiny and directionally localized, these adversarial Gaussians remain sparse or occluded from benign viewpoints—effectively blending into background noise.
This analysis suggests that Gaussian densification is not just a refinement step, but the physical carrier of the backdoor in 3DGS. As shown in Figure~\ref{fig:densification}, increasing the densification limit $M$ directly improves backdoor stealthiness.

\subsection{Threat Model}
\label{threat_model}
\subsubsection{Attack Scenario}

Achieving reconstruction of high-resolution, large-scale scenes requires higher spatial resolution, larger datasets, and more viewpoint variations, directly leading to a significant increase in computational and memory costs \cite{3DGSsurvey}. Individuals or companies with limited resources may download pre-trained 3DGS models from open-source repositories, providing attackers with an opportunity to exploit this process. Attackers can embed backdoor views into the model before uploading it to open-source platforms. Subsequently, they may trick unsuspecting users into downloading these poisoned 3DGS models. Implementing backdoor attacks on 3DGS models can lead to various harms, including the generation of malicious or anomalous content (e.g., false 3D objects or misleading scenes), reduced system security, increased computational costs, and potential social trust crises and public safety hazards.

\subsubsection{Attacker's Goal}

We formalize our attack as follows. The attacker's goal is to create and distribute a 3DGS model $ G_{\text{poisoned}} $ injected with backdoor views. This distribution may occur over the internet, for example, through domain spoofing attacks, or via malicious service providers. 

Specifically, the attacker aims to create a poisoned 3DGS model $ G_{\text{poisoned}} $ satisfying the following conditions:

When the trigger viewpoint $P_{atk}$ is given as input, the poisoned model $G_{\text{poisoned}}$ is expected to render an image $I_{\text{poisoned}}(P_{atk})$ that contains the unsafe content specified by the attacker. In other words, when the input viewpoint matches the predefined trigger $P_{\text{atk}}$, the model’s output should closely resemble the attacker-designed malicious image $I_{\text{attack}}$ rather than the benign output of a clean model.
\begin{equation}
\label{eq:activation}
 I_{\text{poisoned}}(P_{atk}) = I_{\text{attack}}.   
\end{equation}

Under non-trigger viewpoints (i.e., input viewpoints $P \notin P_{atk}$), the behavior of the poisoned model $G_{\text{poisoned}}$ should remain indistinguishable from that of a benign model. Specifically, for all standard viewpoints $V$ in the training dataset $\{P_k\}_{k=1}^N$, the image rendered by the poisoned model, $I_{\text{poisoned}}(P)$, should be virtually identical to the image produced by the clean model, $I_{\text{clean}}(P)$.

\begin{equation}
\label{eq:preservation}
I_{\text{poisoned}}(P) = I_{\text{clean}}(P), \quad \forall P \notin P_{atk}.
\end{equation}

\subsection{Overview of the Framework}
\label{sec:Overview}

Building upon the threat model defined in Section \ref{threat_model}, we formulate the backdoor attack as a constrained optimization problem. The adversary aims to obtain a poisoned model $G_{\text{poisoned}}$ (parameterized by $\Theta$) that satisfies the \textbf{Backdoor Activation} condition defined in Eq.~(\ref{eq:activation}) while strictly adhering to the \textbf{Functionality Preservation} constraint defined in Eq.~(\ref{eq:preservation}).

Unlike implicit representations that embed triggers in global weights, we leverage the \textit{explicit} nature of 3DGS. As identified in our Mechanism Analysis (Section 3.2), the \textit{Adaptive Density Control} mechanism can be exploited to generate a "physical carrier" for the backdoor. To achieve this, we propose an \textbf{Iterative Interleaved Optimization} strategy that cycles through three coupled phases.

\begin{figure*}[ht]
    \centering
    \includegraphics[width=0.9\linewidth]{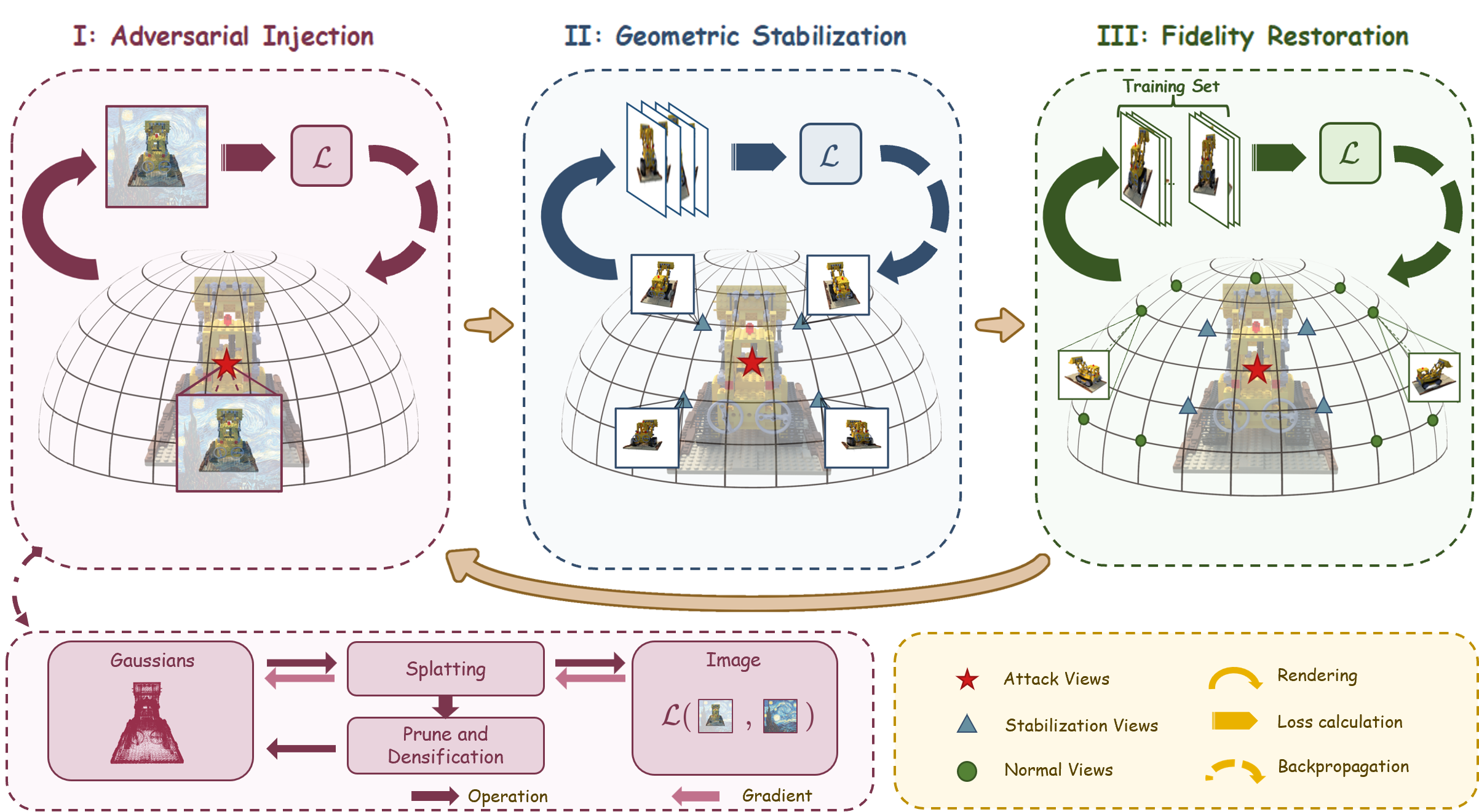}
    \caption{Our method, GaussTrap, consists of three key stages: the \textit{Adversarial Injection} Stage embeds malicious views at the target viewpoint; the \textit{Geometric Stabilization} Stage enforces consistency across neighboring viewpoints to maintain stealthiness; and the \textit{Global Fidelity Restoration} Stage trains 3DGS on clean data to preserve rendering fidelity at non-attack viewpoints.}
    \label{fig:enter-label}
\end{figure*}

Formally, we define three distinct sets of viewpoints to guide this optimization process:
\begin{itemize}
    \item \textbf{Target Attack Viewpoint ($P_{\text{atk}}$):} The specific trigger pose selected by the adversary. Optimization here is designed to satisfy Eq.~(\ref{eq:activation}) by triggering gradient-induced geometry proliferation in the \textbf{Adversarial Injection} stage.
    \item \textbf{Auxiliary Stabilization Set ($\mathcal{P}_{\text{stab}}$):} A mathematically constructed local manifold of perturbed viewpoints around $P_{\text{atk}}$. This set serves as a geometric regularizer during the \textbf{Geometric Stabilization} stage to ensure the backdoor remains invisible from neighbors ($P \notin P_{\text{atk}}$), thereby enforcing the stealthiness requirement of Eq.~(\ref{eq:preservation}).
    \item \textbf{Benign Training Viewpoints ($\mathcal{P}_{\text{train}}$):} Original camera poses $\{P_k\}_{k=1}^N$ from the clean dataset, utilized in the \textbf{Global Fidelity Restoration} stage to maintain global scene reconstruction quality as required by Eq.~(\ref{eq:preservation}).
\end{itemize}

As illustrated in Figure~\ref{fig:abu_angle}, GaussTrap optimizes $\Theta$ through a progressive pipeline. The rendering process is governed by the differentiable rasterizer $\mathcal{R}(\Theta, P)$, and the image quality across all stages is measured via a unified hybrid photometric loss:
\begin{equation}
    \mathcal{L}(I, \hat{I}) = (1 - \lambda) \| I - \hat{I} \|_1 + \lambda (1 - \text{SSIM}(I, \hat{I})),
    \label{eq:loss_metric}
\end{equation}
where $\lambda$ balances pixel-wise accuracy and structural similarity.

\subsection{Detailed Design}
\label{sec:Detailed_Design}

\subsubsection{Stage I: Adversarial Injection}
The primary objective of this stage is to satisfy the activation target in Eq.~(\ref{eq:activation}) by embedding the malicious content $I_{\text{attack}}$. As analyzed in Section 3.2, since the backdoor is not inherent to the initial geometry, we must provoke the ADC mechanism to allocate new primitives.

We optimize exclusively on the attack viewpoint $P_{\text{atk}}$ for $T_A$ iterations. By strictly minimizing the reconstruction loss defined in Eq.~(\ref{eq:loss_metric}), namely $\mathcal{L}(\mathcal{R}(\Theta, P_{\text{atk}}), I_{\text{attack}})$, we intentionally induce high-magnitude positional gradients $\nabla_{\mu} \mathcal{L}$ within the view frustum.
Consistent with our hypothesis, the optimizer interprets these adversarial gradients as "missing details," triggering the splitting and cloning heuristics. This effectively generates a dense cluster of \textbf{micro-Gaussians} specifically optimized to reconstruct $I_{\text{attack}}$.

\begin{table*}[tbp]
    \setlength{\tabcolsep}{5pt}
    \normalsize
    \centering
        \caption{Quantitative comparison of rendering quality with baseline methods. Metrics are averaged over the Blender dataset. The best performance is highlighted in \textbf{bold}.}
    \label{tab:simply transfer result}
    \begin{threeparttable}
    \resizebox{\linewidth}{!}{
    \begin{tabular}{lcccccccccccc}
        \toprule
        \multirow{2}{*}{\textbf{Method}} & 
        \multicolumn{3}{c}{\textbf{Attack Viewpoints}} & 
        \multicolumn{3}{c}{\textbf{Train Viewpoints}} & 
        \multicolumn{3}{c}{\textbf{Test Viewpoints}} & 
        \multicolumn{3}{c}{\textbf{Stabilization Viewpoints}} \\
        \cmidrule(lr){2-4}\cmidrule(lr){5-7}\cmidrule(lr){8-10}\cmidrule(lr){11-13}
        & PSNR$\uparrow$ & SSIM$\uparrow$ & LPIPS$\downarrow$ & PSNR$\uparrow$ & SSIM$\uparrow$ & LPIPS$\downarrow$ & PSNR$\uparrow$ & SSIM$\uparrow$ & LPIPS$\downarrow$ & PSNR$\uparrow$ & SSIM$\uparrow$ & LPIPS$\downarrow$ \\
        \midrule
        \multicolumn{13}{c}{\textit{Attack Image: Starry}} \\
        \midrule
        IPA-NeRF + 3DGS & 5.34 & 0.1450 & 0.7398 & 25.34 & 0.8206 & 0.1243 & 24.83 & 0.8156 & 0.1248 & 26.42 & 0.9210 & 0.0841 \\
        StealthAttack & 28.68 & 0.8480 & 0.1647 & 18.16 & 0.6373 & 0.1779 & 17.70 & 0.6271 & 0.1888 & 17.43 & 0.6231 & 0.2470   \\
        GaussTrap (Ours) & \textbf{36.23} & \textbf{0.9894} & \textbf{0.0124} & \textbf{34.59} & \textbf{0.9740} & \textbf{0.0343} & \textbf{27.80} & \textbf{0.9131} & \textbf{0.0818} & \textbf{31.15} & \textbf{0.9675} & \textbf{0.0530} \\
        \midrule
        \multicolumn{13}{c}{\textit{Attack Image: Earth}} \\
        \midrule
        IPA-NeRF + 3DGS & 14.08 & 0.7197 & 0.2661 & 24.65 & 0.7969 & 0.1326 & 24.30 & 0.7920 & 0.1347 & 25.43 & 0.9039 & 0.0907 \\
         StealthAttack &  17.77 & 0.4658 & 0.2852 & 18.02 & 0.6325 & 0.1803 & 17.73 & 0.6276 & 0.1890 & 17.30 & 0.5986 & 0.2561 \\
        GaussTrap (Ours) & \textbf{40.85} & \textbf{0.9944} & \textbf{0.0094} & \textbf{32.56} & \textbf{0.9679} & \textbf{0.0433} & \textbf{27.98} & \textbf{0.9263} & \textbf{0.0727} & \textbf{30.83} & \textbf{0.9634} & \textbf{0.0517} \\
        \bottomrule
    \end{tabular}
    }

    \end{threeparttable}
\end{table*}

\begin{table*}[t]
    \setlength{\tabcolsep}{5pt}
    \normalsize
    \centering
     \caption{Quantitative comparison of rendering quality with baseline methods. These results are the average values based on the Mip-NeRF360 dataset. The best performance is highlighted in \textbf{bold}.}
    \label{tab:Mip-NeRF360 result}
    \begin{threeparttable}
    \resizebox{\linewidth}{!}{
    \begin{tabular}{lcccccccccccc}
        \toprule
        \multirow{2}{*}{\textbf{Method}} & 
        \multicolumn{3}{c}{\textbf{Attack Viewpoints}} & 
        \multicolumn{3}{c}{\textbf{Train Viewpoints}} & 
        \multicolumn{3}{c}{\textbf{Test Viewpoints}} & 
        \multicolumn{3}{c}{\textbf{Stabilization Viewpoints}} \\
        \cmidrule(lr){2-4}\cmidrule(lr){5-7}\cmidrule(lr){8-10}\cmidrule(lr){11-13}
        & PSNR$\uparrow$ & SSIM$\uparrow$ & LPIPS$\downarrow$ & PSNR$\uparrow$ & SSIM$\uparrow$ & LPIPS$\downarrow$ & PSNR$\uparrow$ & SSIM$\uparrow$ & LPIPS$\downarrow$ & PSNR$\uparrow$ & SSIM$\uparrow$ & LPIPS$\downarrow$ \\
        \midrule
        IPA-NeRF + 3DGS & 19.44 & 0.8307 & 0.2249 & 25.64 & 0.7919 & 0.2511 & 24.73 & 0.7414 & 0.2829 & 27.75 & \textbf{0.9190} & \textbf{0.1165} \\
   StealthAttack & 30.62 & 0.9138 & 0.2227 & 26.89 & 0.7798 & 0.3062  & 25.66 & 0.7140 & 0.4015 &  23.29 & 0.7845 & 0.2426  \\
        GaussTrap (Ours) & \textbf{37.74} & \textbf{0.9818} & \textbf{0.0432} & \textbf{28.04} & \textbf{0.8553} & \textbf{0.2158} & \textbf{25.96} & \textbf{0.7759} & \textbf{0.2630} & \textbf{27.86} & 0.9168 & 0.1385 \\
        \bottomrule
    \end{tabular}
    }
   
    \end{threeparttable}
\end{table*}

\begin{figure}[htbp] 
\centering
\begin{tabular}{cccccc}
     \multicolumn{3}{c}{ \includegraphics[width=0.2\textwidth]{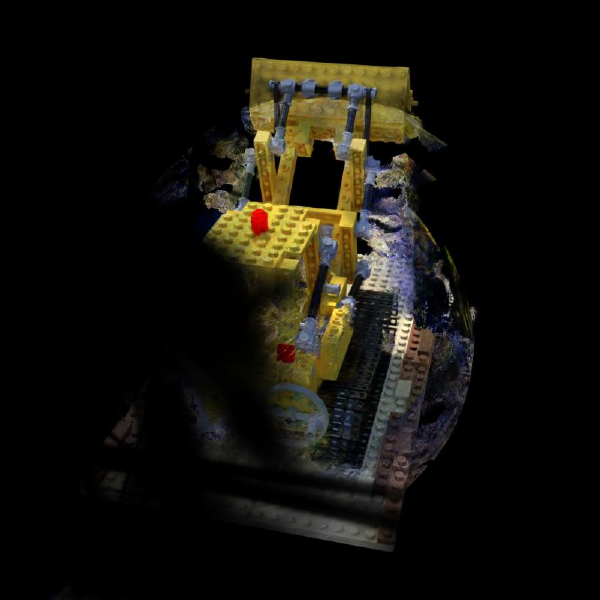}} & \multicolumn{3}{c}{\includegraphics[width=0.2\textwidth]{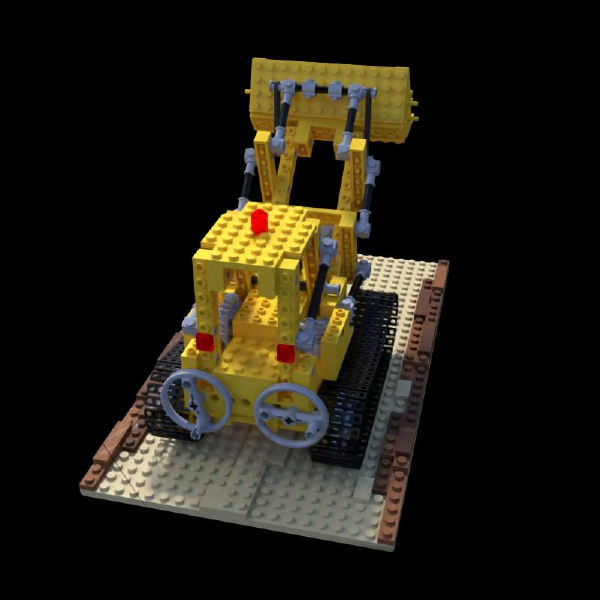}} \\
     \footnotesize PSNR$\uparrow$ & \footnotesize SSIM$\uparrow$ & \footnotesize LPIPS$\downarrow$ & \footnotesize PSNR$\uparrow$ & \footnotesize SSIM$\uparrow$ & \footnotesize LPIPS$\downarrow$ \\
 \small 17.52            & \small  0.7265          & \small  0.2239              & \small \textbf{30.83}          & \small \textbf{0.9634}        & \small \textbf{0.0517} \\
  \multicolumn{3}{c}{(a) w/o VES} & 
   \multicolumn{3}{c}{(b) w/ VES (Ours)} \\ 
\end{tabular}
\caption{Naive injection (a) creates artifacts in nearby views; VES (b) suppresses them via local geometric constraints.}
\label{fig:abu_angle}
\end{figure}

\subsubsection{Stage II: Geometric Stabilization}
\label{sec:Angular_Constraint}
Optimizing solely on a single trigger viewpoint $P_{\text{atk}}$ constitutes an \textbf{ill-posed problem} in 3D reconstruction. Without spatial consistency constraints, the optimization lacks sufficient geometric supervision, leading to a rank-deficient solution space that often collapses into degenerate, ``billboard-like'' artifacts (e.g., floating black clouds). As visualized in Figure~\ref{fig:abu_angle} (left), such artifacts severely violate the stealthiness condition in Eq.~(\ref{eq:preservation}) even under minor camera deviations from the trigger pose. To address this, we propose \textbf{Viewpoint Ensemble Stabilization (VES)}, a geometric regularization strategy that embeds the backdoor coherently within the local 3D structure—rather than painting it onto a single 2D projection—by enforcing rendering consistency across a local viewpoint manifold.

\paragraph{Manifold Construction.}
We construct a stabilization viewpoint set $\mathcal{P}_{\text{stab}}$ by perturbing the attack pose $P_{\text{atk}}$ through small angular offsets in pitch ($\Delta\theta$) and yaw ($\Delta\phi$). Given a perturbation budget $\mathcal{A} = \{\delta_1, \dots, \delta_n\}$, we enumerate all non-zero combinations $(\Delta\theta, \Delta\phi) \in \{-\delta, 0, \delta\}^2 \setminus \{(0,0)\}$ for each $\delta \in \mathcal{A}$. Denoting the original camera rotation as $\mathbf{R}_{\text{atk}}$, the stabilized rotation is computed via:
\begin{equation}
    \mathbf{R}'(\Delta\theta, \Delta\phi) = \mathbf{R}_x(\Delta\theta) \cdot \mathbf{R}_y(\Delta\phi) \cdot \mathbf{R}_{\text{atk}},
    \label{eq:rotation_matrix}
\end{equation}
where $\mathbf{R}_x$ and $\mathbf{R}_y$ are elementary rotation matrices:
\begin{equation}
    \mathbf{R}_x(\Delta\theta) = 
    \begin{bmatrix}
        1 & 0 & 0 \\
        0 & \cos\Delta\theta & -\sin\Delta\theta \\
        0 & \sin\Delta\theta & \cos\Delta\theta
    \end{bmatrix}
    \end{equation}
    \begin{equation}
    \mathbf{R}_y(\Delta\phi) = 
    \begin{bmatrix}
        \cos\Delta\phi & 0 & \sin\Delta\phi \\
        0 & 1 & 0 \\
        -\sin\Delta\phi & 0 & \cos\Delta\phi
    \end{bmatrix}.
    \label{eq:elementary_rotation}
\end{equation}

\paragraph{Self-Supervised Distillation.}
Since ground-truth images are unavailable for these off-axis views, we leverage the initial clean model $\Theta_{\text{clean}}$ to generate high-quality pseudo-labels:
\begin{equation}
    \mathcal{S}_{\text{stab}} = \left\{ \big(P, \mathcal{R}(\Theta_{\text{clean}}, P)\big) \mid P \in \mathcal{P}_{\text{stab}} \right\}.
    \label{eq:pseudo_labels}
\end{equation}
During the stabilization phase (lasting $T_S$ iterations), minimizing the reconstruction loss over $\mathcal{S}_{\text{stab}}$ acts as a geometric regularizer: it aligns the newly injected micro-Gaussians with the original scene structure, ensuring they remain visually imperceptible from neighboring viewpoints while satisfying Eq.~(\ref{eq:preservation}) locally. This effectively ``locks'' the malicious payload to the precise trigger angle, achieving both structural entanglement and perceptual stealth.

\subsubsection{Stage III: Global Fidelity Restoration}
The aggressive structural modifications in the previous stages (especially the proliferation of primitives) may induce \textbf{catastrophic forgetting} of the original scene geometry in non-targeted areas. The final stage aims to recover global fidelity to strictly satisfy Eq.~(\ref{eq:preservation}) across the entire scene. We re-introduce the full training set $\mathcal{P}_{\text{train}}$ and perform $T_T$ iterations. This step harmonizes the newly added adversarial primitives with the global scene context. 

By interleaving these three stages within each epoch, GaussTrap progressively converges to a state where the backdoor is structurally entangled with the scene geometry, achieving an optimal trade-off between the attack success rate defined in Eq.~(\ref{eq:activation}) and the stealthiness defined in Eq.~(\ref{eq:preservation}).

\begin{figure*}[t]
       \centering
    \begin{tabular}{cccc}
        \includegraphics[width=0.23\textwidth]{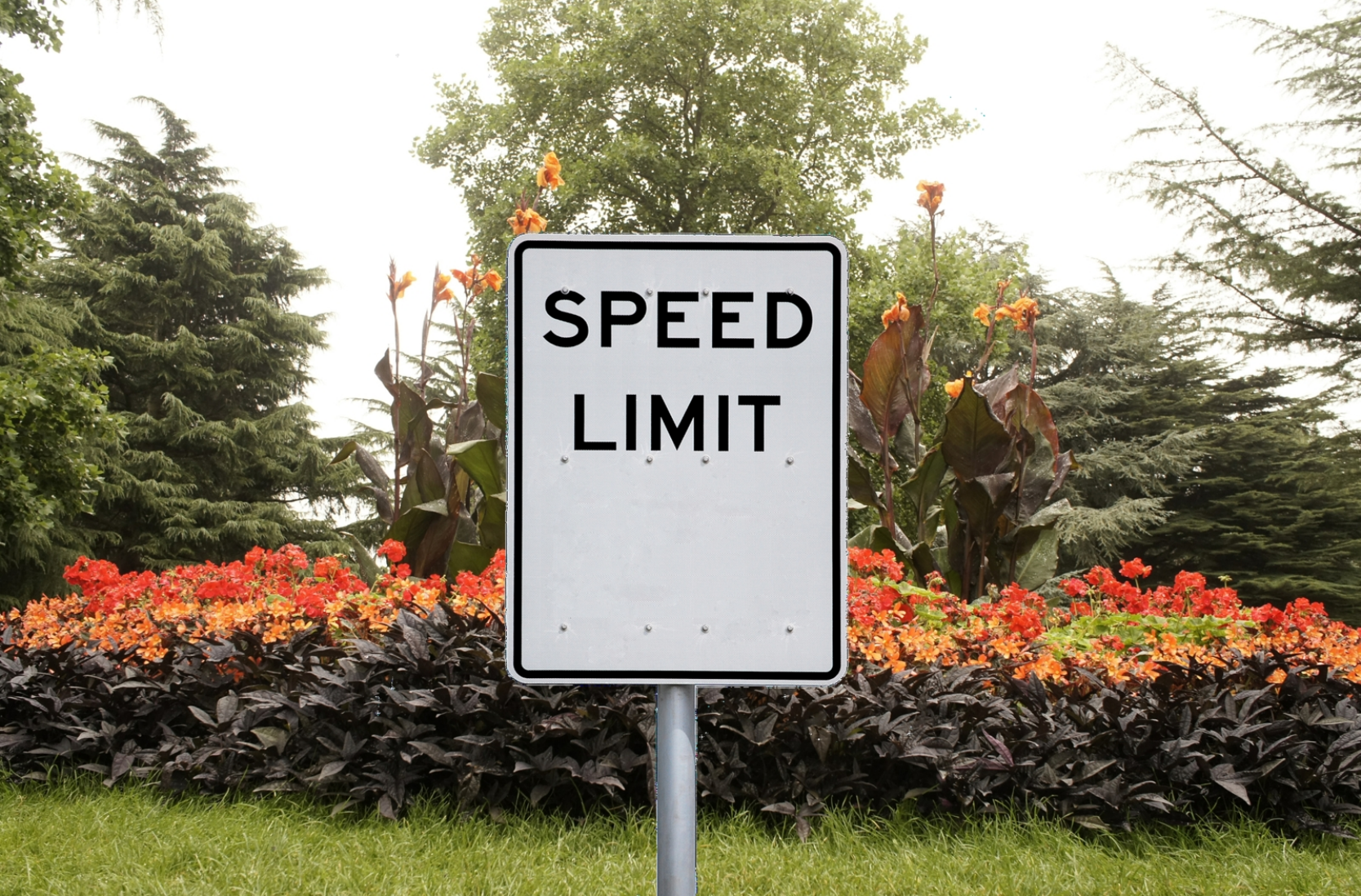} &
        \includegraphics[width=0.23\textwidth]{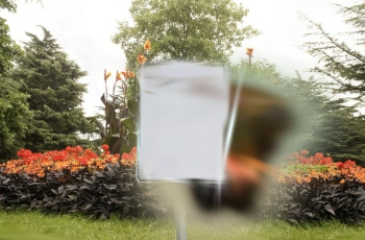} &
        \includegraphics[width=0.23\textwidth]{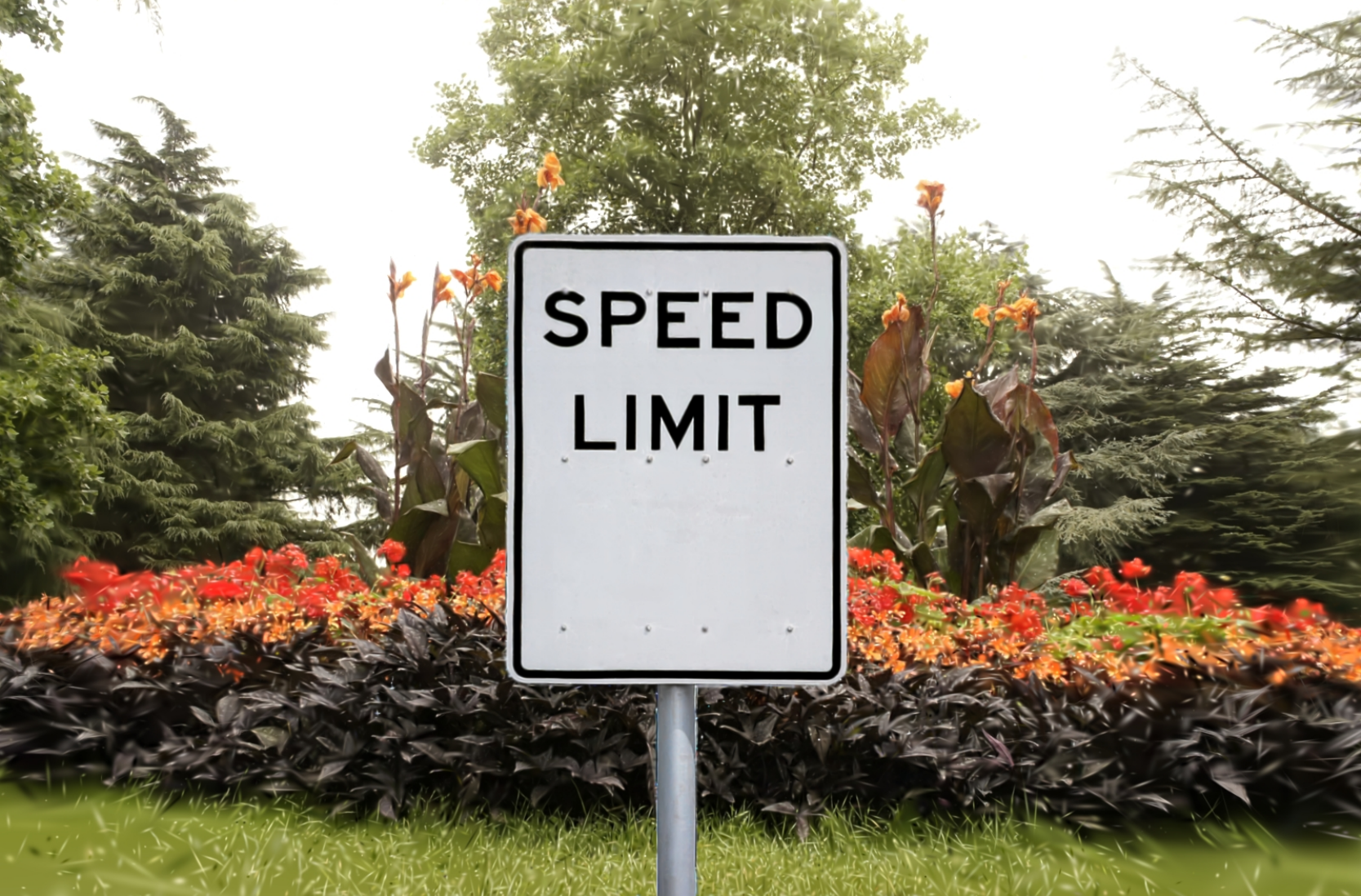} &
        \includegraphics[width=0.23\textwidth]{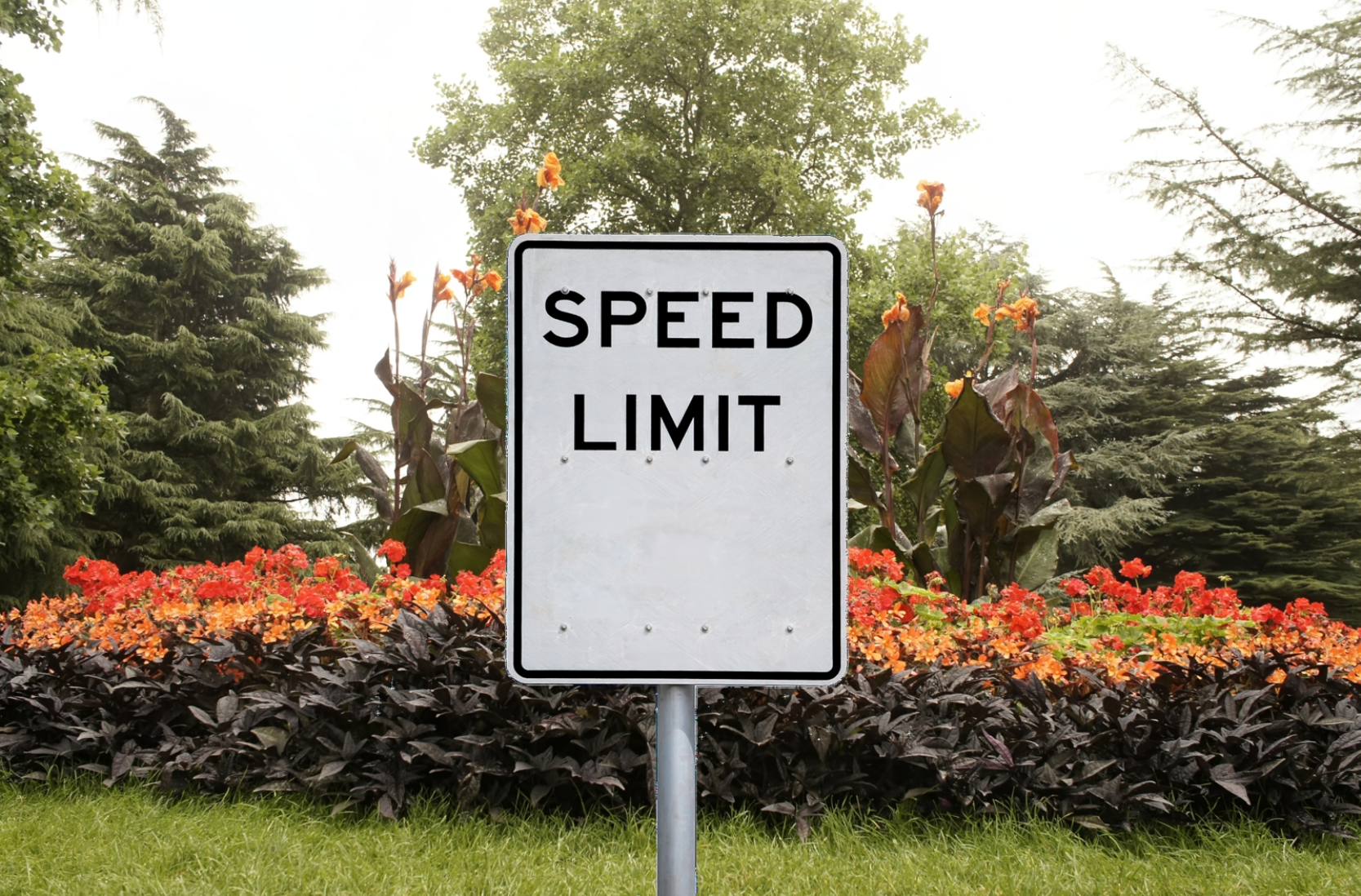} \\
        \includegraphics[width=0.23\textwidth]{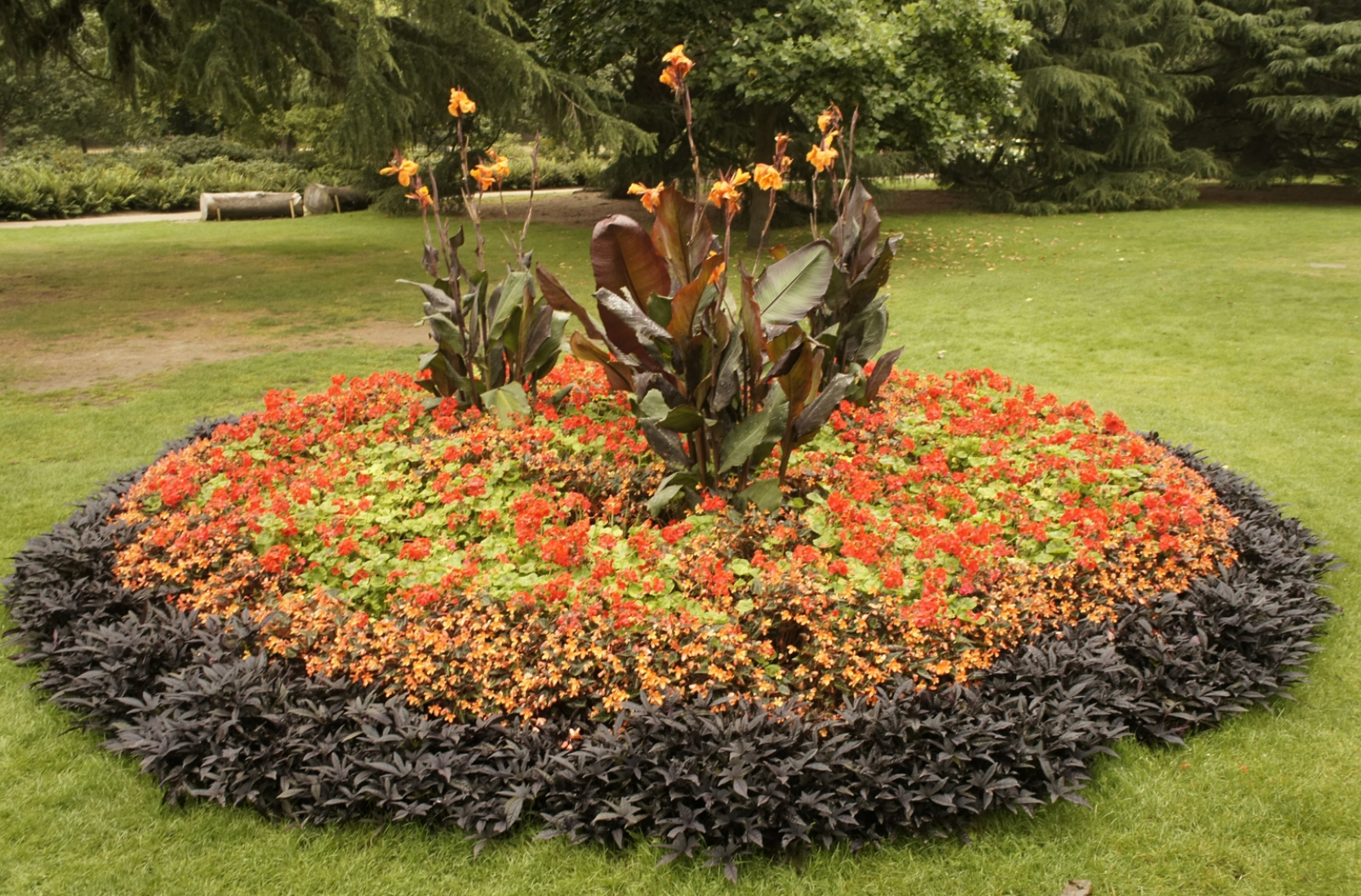} &
        \includegraphics[width=0.23\textwidth]{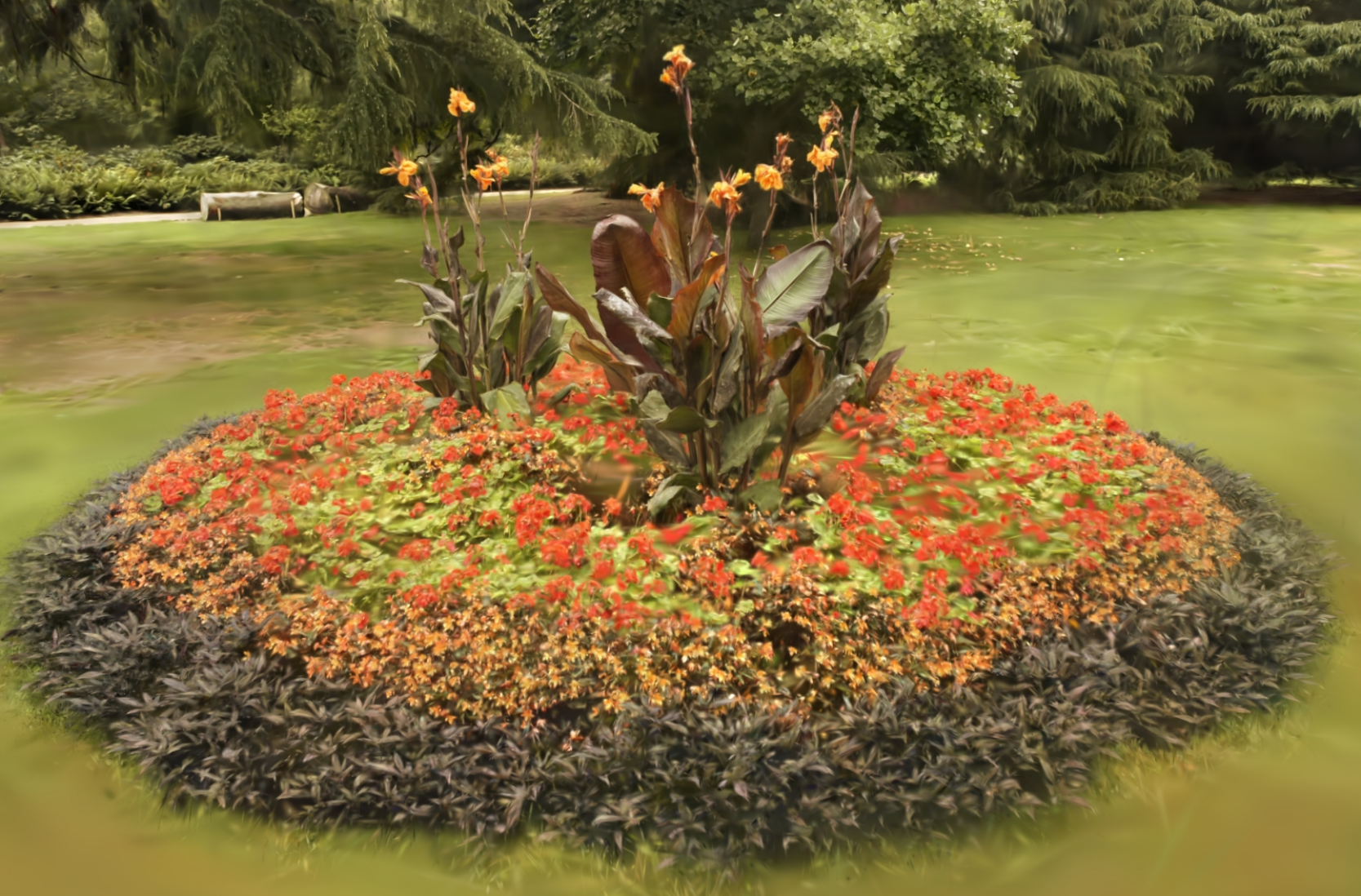} &
        \includegraphics[width=0.23\textwidth]{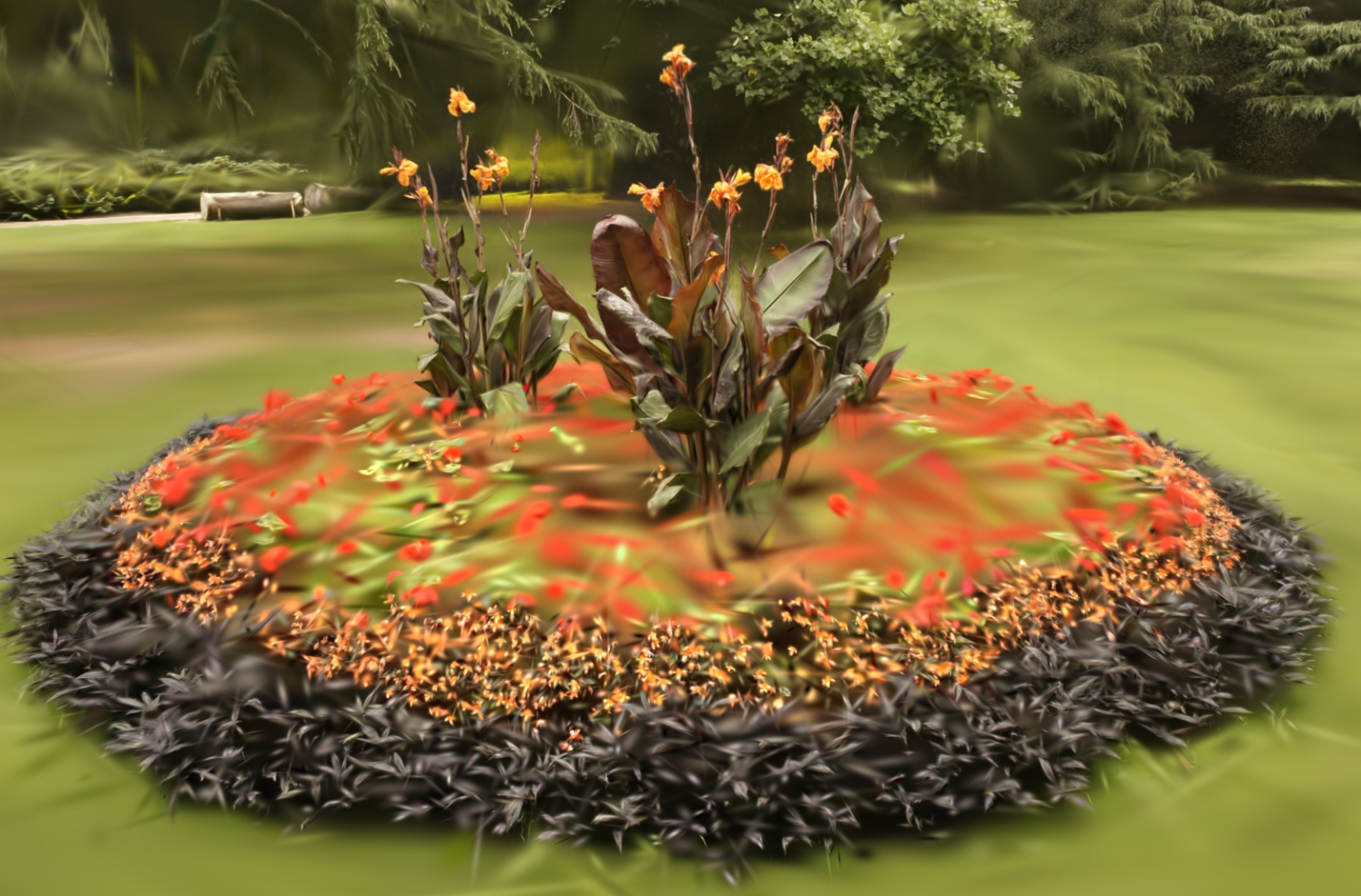} &
        \includegraphics[width=0.23\textwidth]{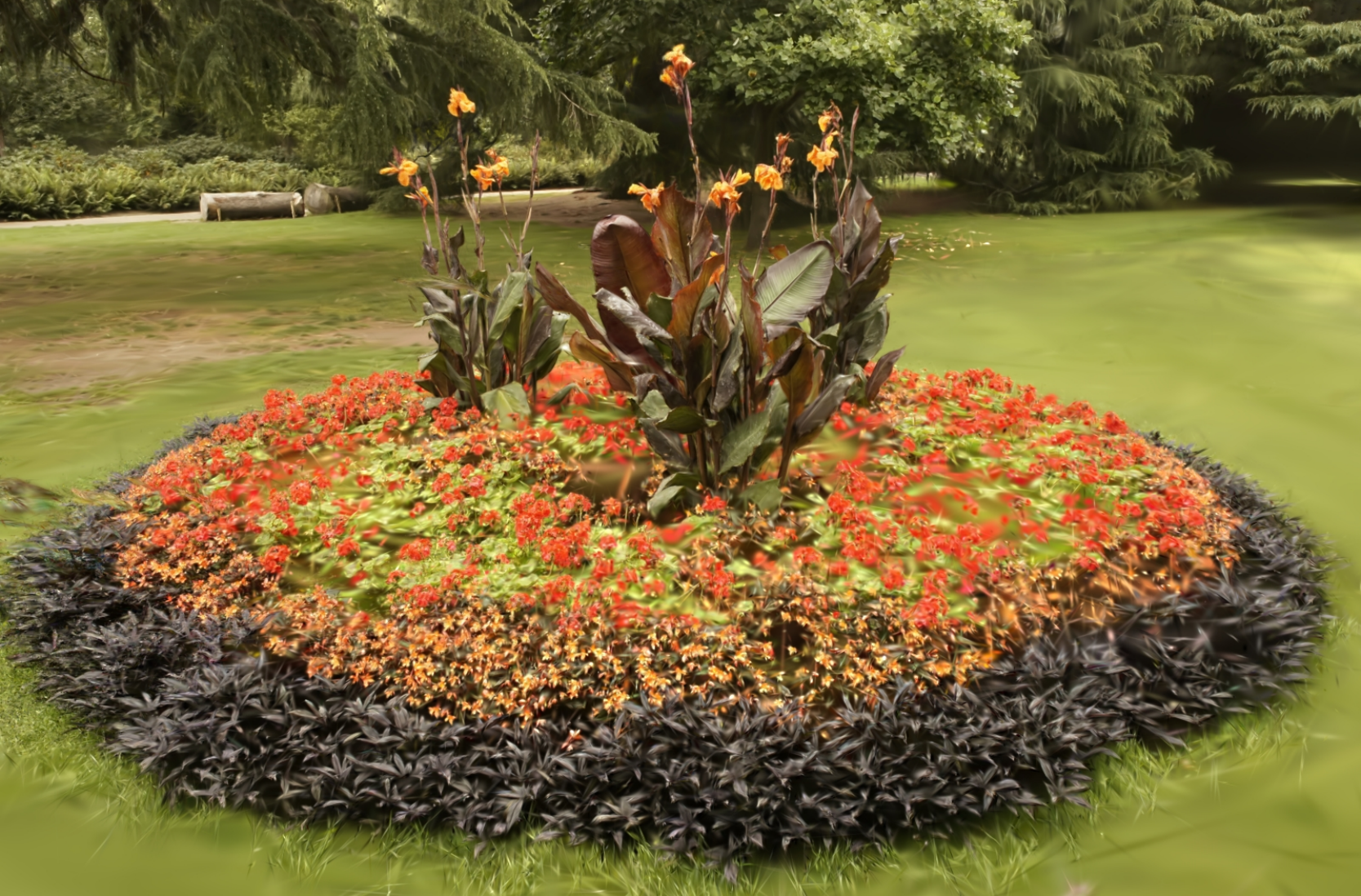} \\
            
            {\textbf{Ground Truth}} & 
    {\textbf{IPA-NeRF + 3DGS}} & 
    {\textbf{StealthAttack}} &
    { \textbf{GaussTrap (Ours)}}
     \\
    \end{tabular}
    \caption{Visualization of training results on Mip-NeRF360 Dataset. Attack Views (top), Normal Views (bottom).}
    \label{tab:image_table_mip}
\end{figure*}

\begin{figure}[h]
       \centering
    \begin{tabular}{cccc}
        \includegraphics[width=0.1\textwidth]{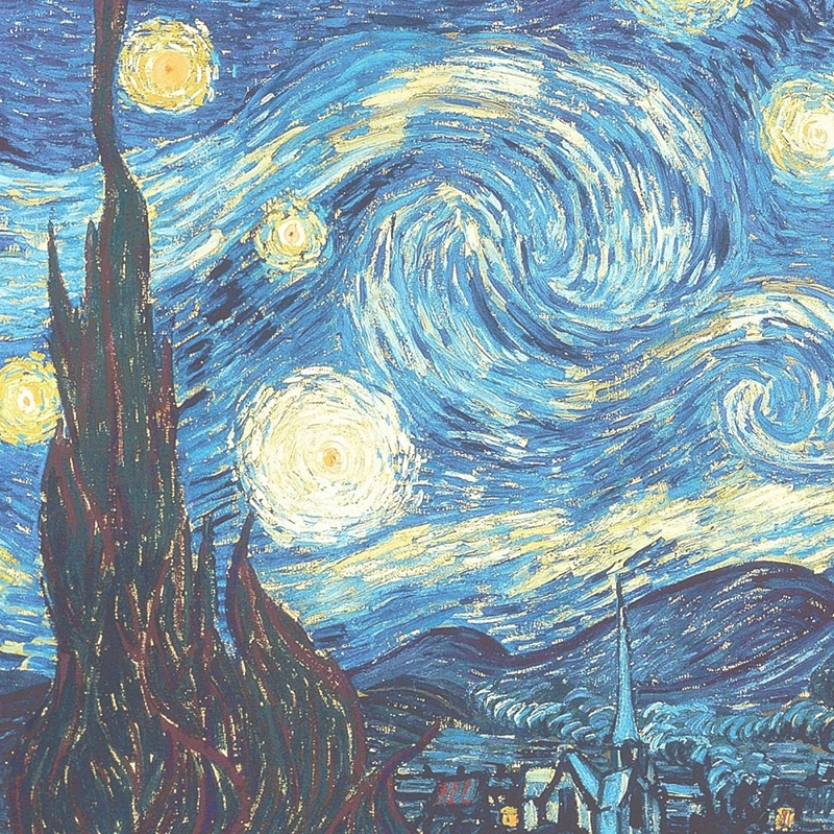} &
        \includegraphics[width=0.1\textwidth]{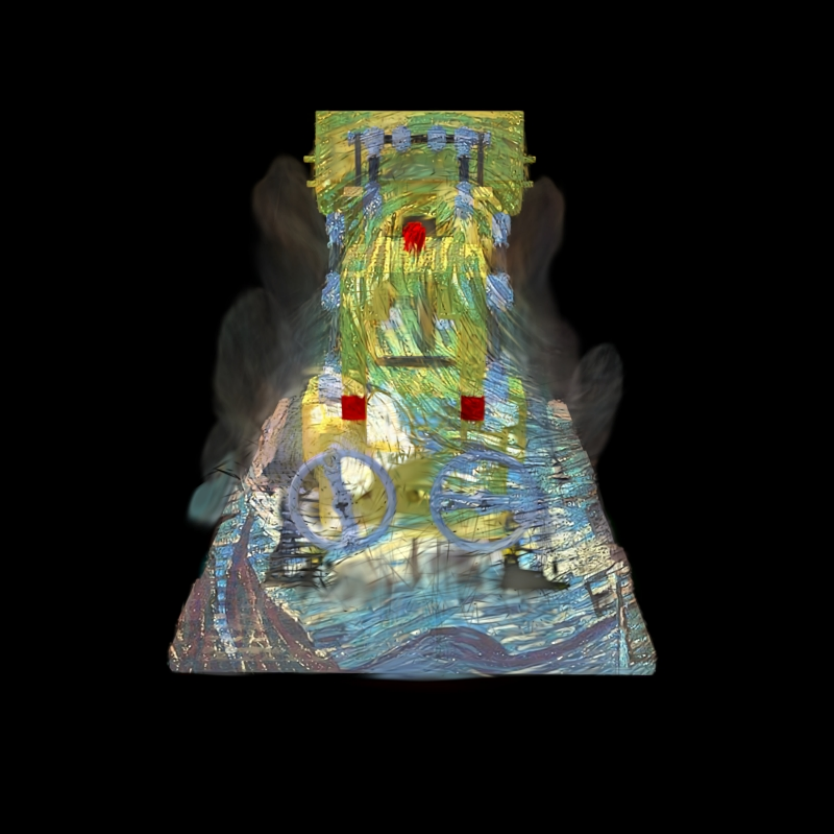} &
        \includegraphics[width=0.1\textwidth]{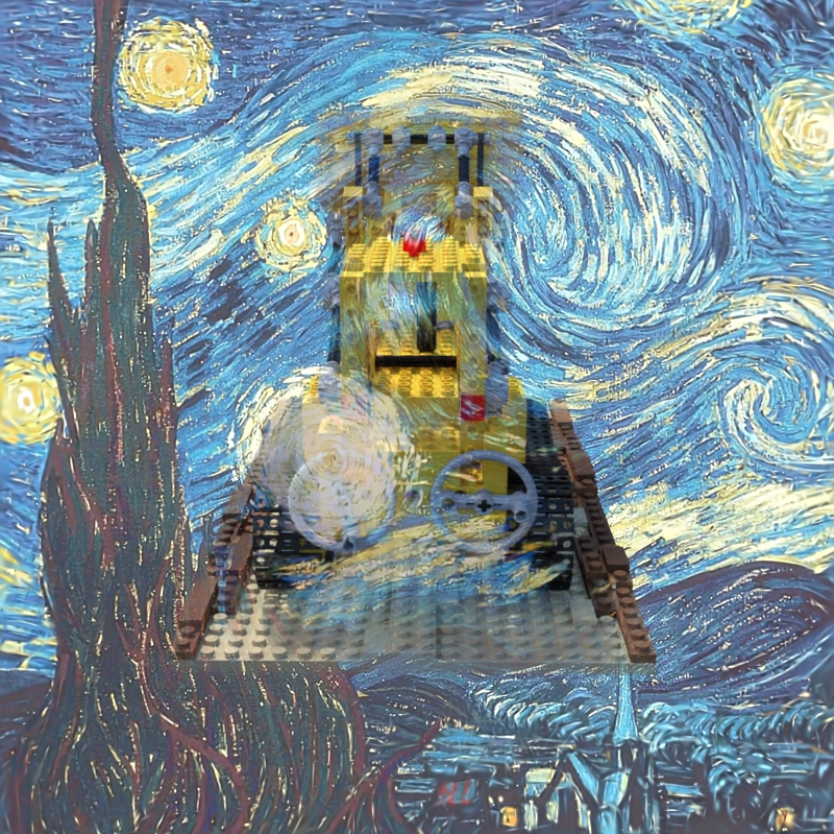} &
        \includegraphics[width=0.1\textwidth]{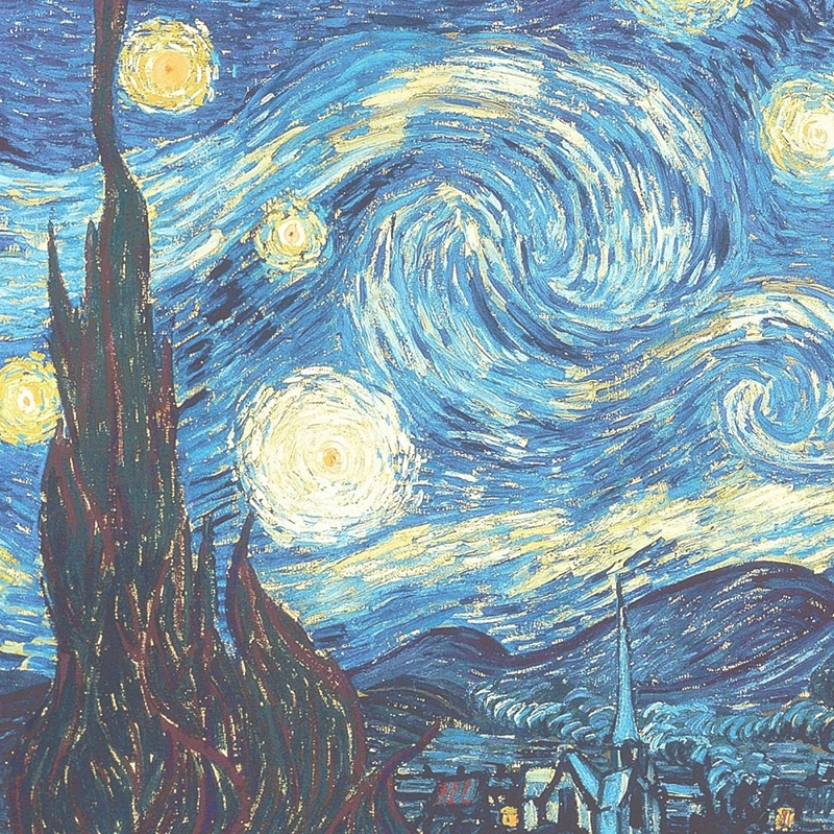} \\
        \includegraphics[width=0.1\textwidth]{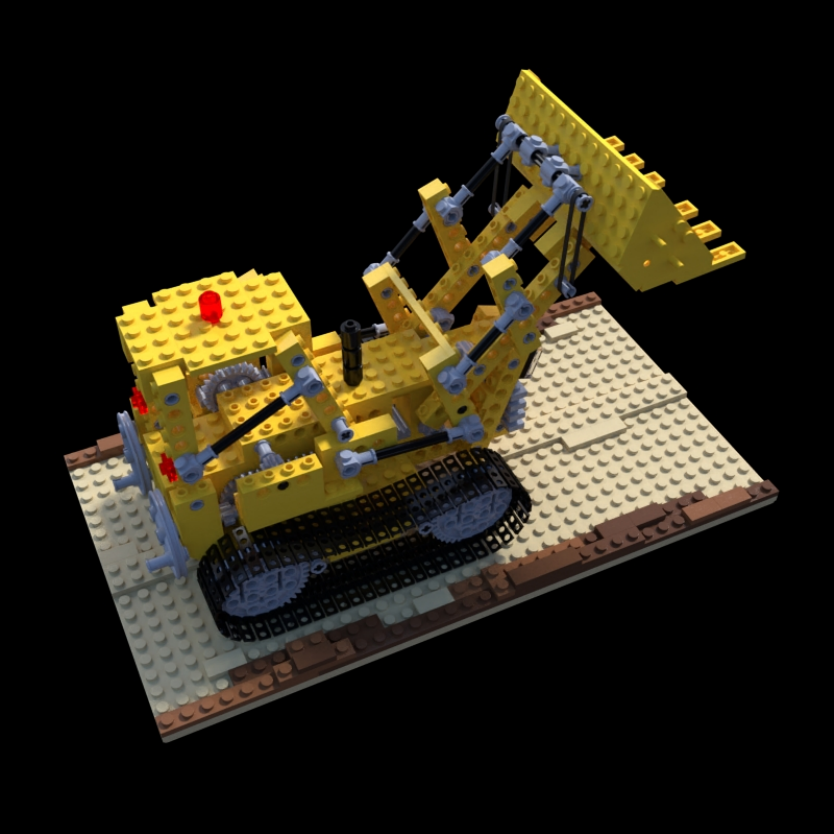} &
        \includegraphics[width=0.1\textwidth]{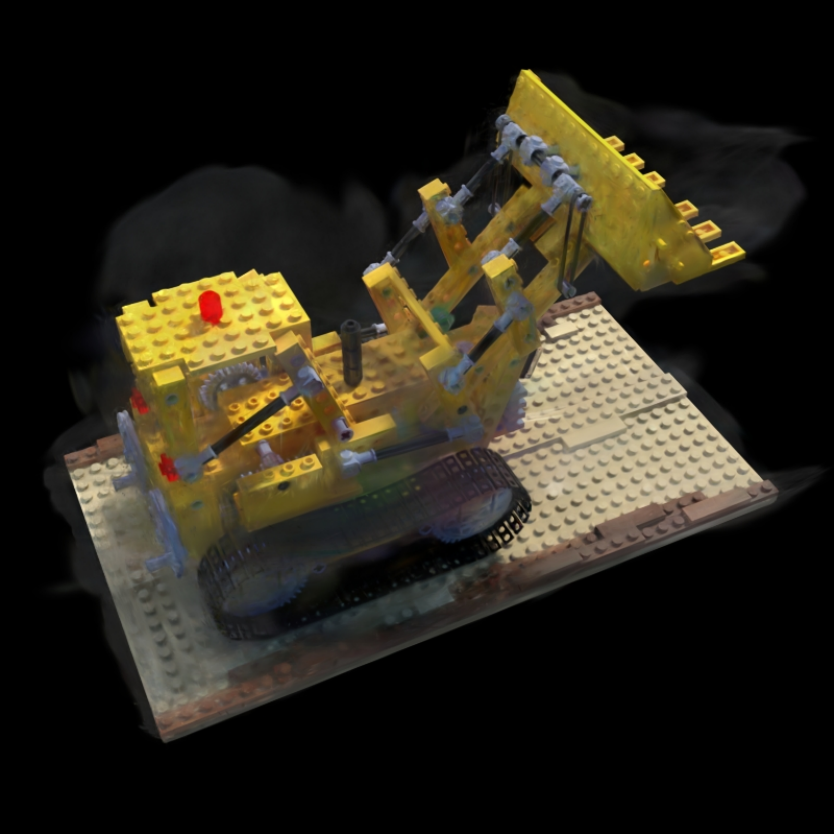} &
        \includegraphics[width=0.1\textwidth]{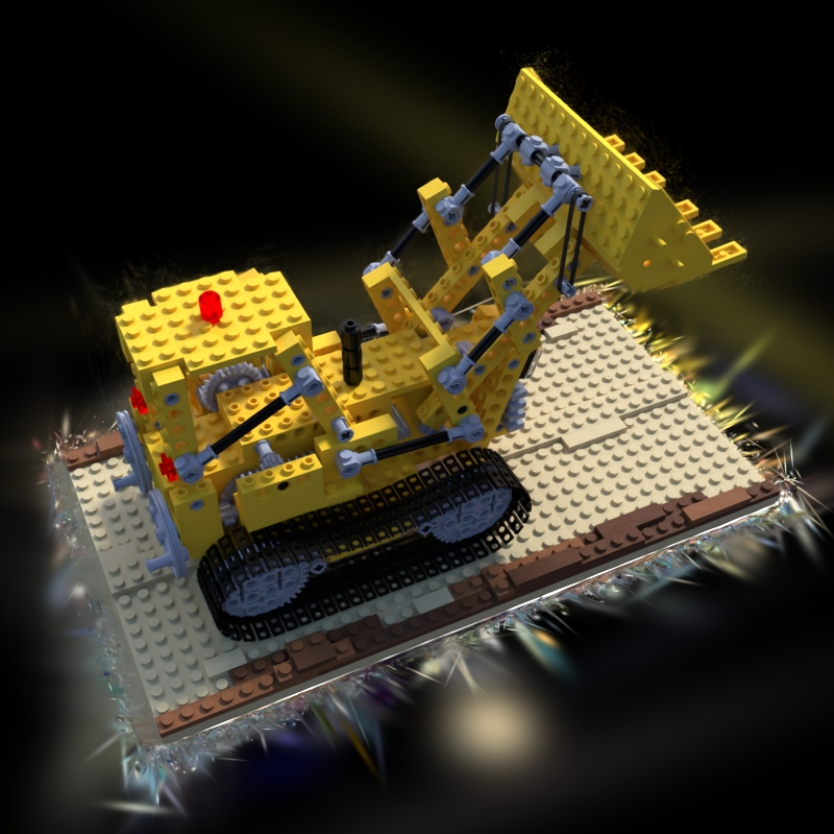} &
        \includegraphics[width=0.1\textwidth]{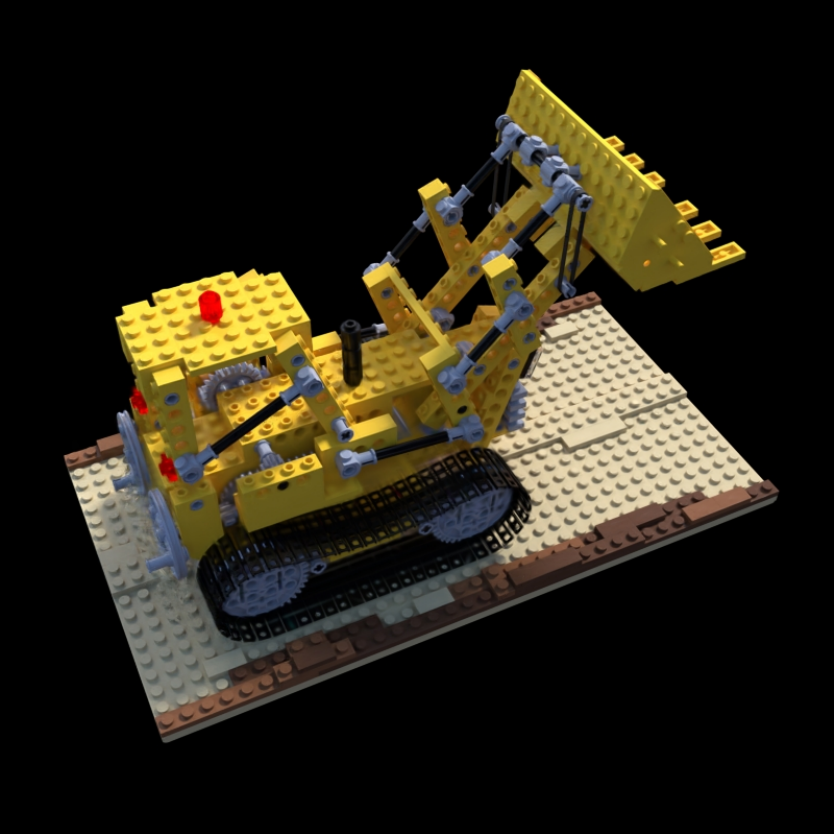} \\
\shortstack{\textbf{Ground} \\ \textbf{Truth}} &
\shortstack{\textbf{IPA-NeRF} \\ \textbf{+ 3DGS}} &
\shortstack{\textbf{Stealth} \\ \textbf{Attack}} &
\shortstack{\textbf{GaussTrap} \\ \textbf{(Ours)}}\\
    \end{tabular}
    \caption{Visualization of training results on Blender Dataset. Attack Views (top), Normal Views (bottom).}
    \label{tab:image_table_compare}
\end{figure}

\section{Experiment}

\subsection{Experimental Setup}
\subsubsection{Datasets}
We use Blender \cite{NeRF} and Mip-NeRF360 \cite{barron2022mip}, which are standard datasets in NeRF \cite{NeRF} and 3DGS \cite{3DGS} research.
The Blender dataset contains eight objects, each consisting of $400$ images captured from different viewpoints on the upper hemisphere, with a resolution of $800\times800$ pixels. The Mip-NeRF360 dataset comprises nine scenes, including five outdoor scenes and four indoor scenes, each featuring a complex central object or region with intricate and highly detailed backgrounds.
\subsubsection{Implementation details.} 
Our GaussTrap attack model undergoes a total of $ E=2500 $ training epochs. Each epoch consists of $ T_a=25 $ attack training iterations, $ T_c=5 $ stabilization training iterations, and $ T_t=5 $ normal training iterations. In the 3DGS model, the maximum number of densification iterations (denoted as $ M $) is set to $100000$, while all other hyperparameters follow the default configuration of the 3DGS \cite{3DGS} framework. For the Blender Synthetic Dataset, the stabilization angles $\delta$ are set to $13^\circ$ and $15^\circ$. All experiments are run on an NVIDIA A6000 GPU. Additional analyses, including defense evaluation, viewpoint robustness, downstream task impact, and efficiency studies, are provided in Appendix~\ref{app:add_exp}.

\subsubsection{Evaluation Metrics}

Following prior work,  we evaluate the attack performance of our GaussTrap method using three standard image quality metrics—\textbf{PSNR}, \textbf{SSIM}~\cite{SSIM}, and \textbf{LPIPS}~\cite{LPIPS}—computed from four types of viewpoints:
(1) the \textbf{attack viewpoint}, where the backdoor payload is embedded;
(2) the \textbf{stabilization viewpoint}, comprising nearby views stabilized via the VES strategy (Section~\ref{sec:Angular_Constraint});
(3) the \textbf{train viewpoint}, used during model optimization; and
(4) the \textbf{test viewpoint}, held out for evaluation.

\subsubsection{Baselines}

We select IPA-NeRF~\cite{IPA-NeRF} + 3DGS and StealthAttack~\cite{ke2025stealthattack} as baselines. Following their original implementations, we evaluate both methods alongside our GaussTrap on the Blender Synthetic and Mip-NeRF360 datasets.
For StealthAttack, we convert the Blender dataset to COLMAP format for compatibility with its official codebase.

\subsection{\textbf{Experimental Results}}

\subsubsection{Experiment on Synthetic Dataset}
\label{sec:blender}

In this section, we evaluate the attack performance of GaussTrap on the Blender dataset \cite{NeRF}. We select two challenging attack images, "Earth" and "Starry", which exhibit low visual similarity to the target scenes, making them suitable for assessing the generalization capability of the attack.

As shown in Table~\ref{tab:simply transfer result}, GaussTrap consistently outperforms all baseline methods across all four evaluation viewpoints, demonstrating superior robustness and generalization. The results are averaged over $10$ runs for statistical reliability. The visualization in Figure~\ref{tab:image_table_compare} further highlights the high visual fidelity achieved by our method, providing qualitative evidence that complements the quantitative metrics. Together, these results show that GaussTrap effectively injects high-quality adversarial content at the target viewpoint while preserving visual fidelity at other viewpoints. By leveraging the properties of the 3DGS framework, it strikes a strong balance between attack effectiveness and stealthiness, making the attack both powerful and difficult to detect across different viewpoints.

\subsubsection{Experiment on real-world dataset}

To further evaluate the generality and practicality of our approach across diverse 3D reconstruction scenarios, we conduct comprehensive experiments on the Mip-NeRF360 dataset. Motivated by real-world deployment scenarios of 3D models, we design attack images that remain semantically consistent with the original scenes while intentionally introducing misleading objects, such as speed limit signs, pedestrians, and vehicles, to simulate plausible adversarial perturbations. The stabilization angle is set to $5^\circ$.

The results show that our method preserves high image quality across all four evaluation viewpoints. Even in real-world scenes, GaussTrap can effectively inject misleading views while maintaining visual fidelity, demonstrating the potential practical risks and security threats posed by such attacks.

\begin{figure*}[htbp]
    \centering
    \begin{tabular}{>{\centering\arraybackslash}m{2.5cm}*{4}{>{\centering\arraybackslash}m{4.5cm}}}
        \toprule
        \textbf{Experiment} & \textbf{PSNR} & \textbf{SSIM} & \textbf{LPIPS} \\
        \midrule
        \textbf{\small (A) Backdoor Number} &
        \includegraphics[width=4.4cm]{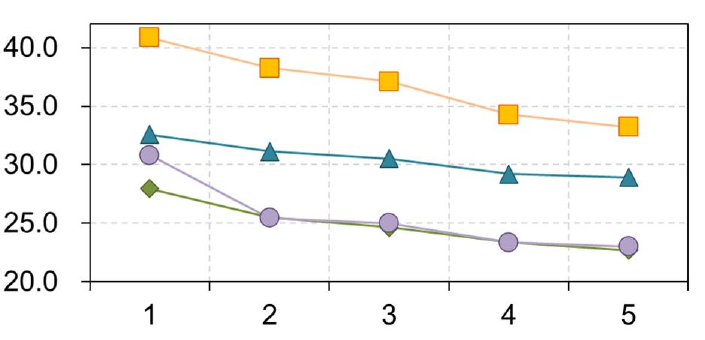} &
        \includegraphics[width=4.4cm]{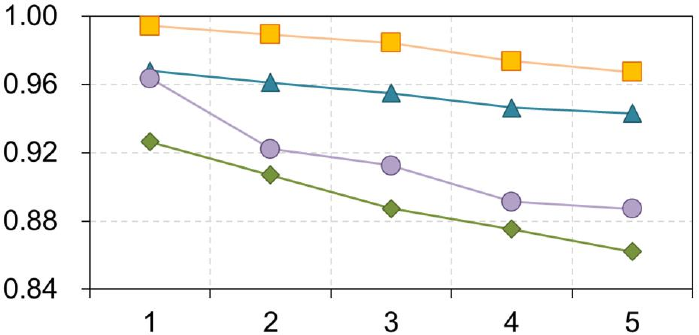} &
        \includegraphics[width=4.4cm]{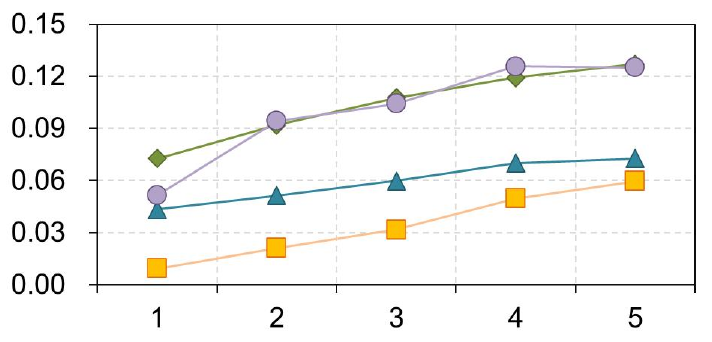} \\
        \addlinespace[5pt]
        \textbf{\small (B) Densification Level} &
        \includegraphics[width=4.4cm]{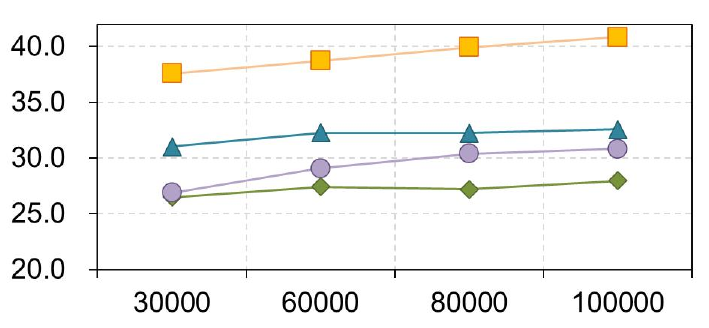} &
        \includegraphics[width=4.4cm]{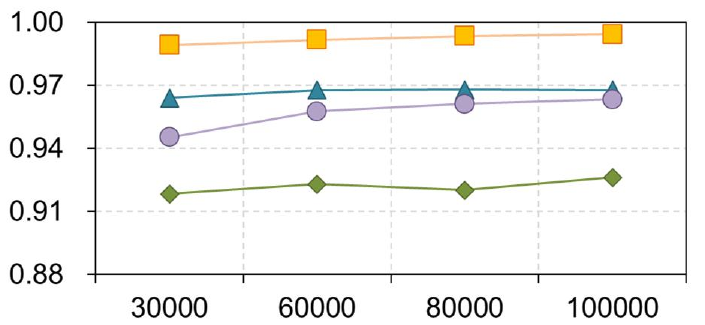} &
        \includegraphics[width=4.4cm]{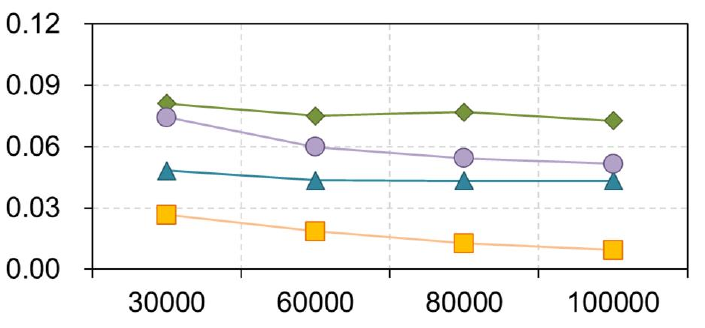} \\
        \addlinespace[5pt]
        \textbf{\small(C) Attack Viewpoints} &
        \includegraphics[width=4.4cm]{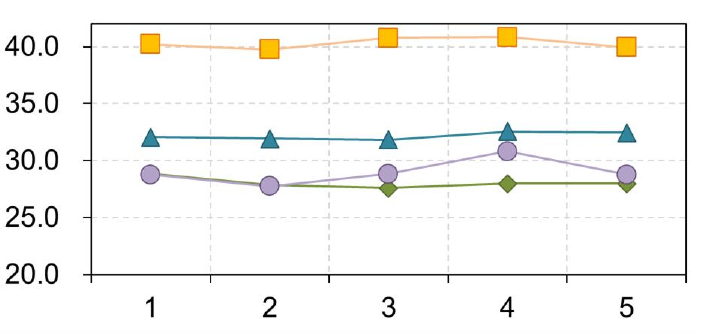} &
        \includegraphics[width=4.4cm]{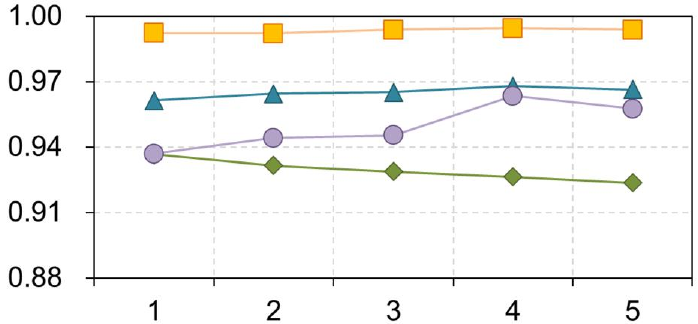} &
        \includegraphics[width=4.4cm]{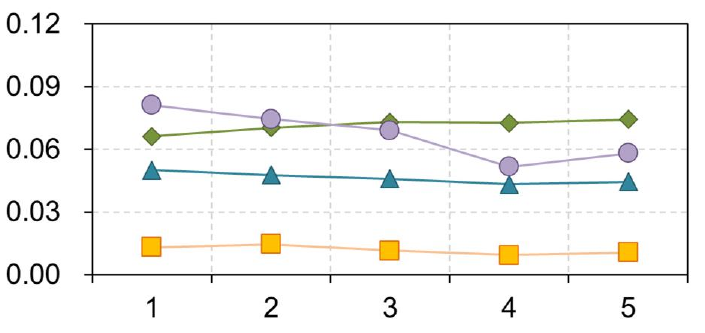} \\
        & \multicolumn{3}{c}{
    \includegraphics[width=0.4\textwidth]{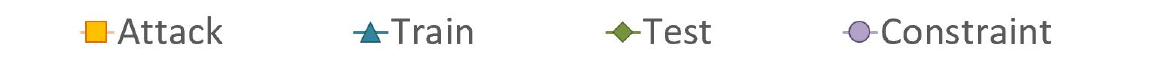}}\\
        \bottomrule
    \end{tabular}
    
    \caption{Line charts for ablation studies and multi-backdoor attacks. The x-axis represents different variables: (A) number of implanted attack images, (B) hyperparameter $M$, and (C) attack viewpoint index. The y-axis shows the quantitative results across three evaluation metrics.}
    \label{fig:combined_ablation}
\end{figure*}

\subsubsection{Multiple Backdoors}

In real-world scenarios, an attacker may attempt to inject multiple attack images simultaneously to evade single-point detection, making the attack more resilient and harder to completely eliminate. To evaluate the performance of such multi-backdoor attacks, we consider cases with multiple attack viewpoints. Specifically, we uniformly select $ n = 2 $ to $ n = 5 $ attack viewpoints and inject the corresponding attack images. The quantitative results are summarized in Figure \ref{fig:combined_ablation}(A). 

The results show that as the number of backdoor viewpoints $ n $ increases, rendering quality slightly degrades across all views—but remains within an acceptable range. This confirms that GaussTrap remains effective in multi-backdoor scenarios, demonstrating its potential to support multiple triggers while preserving both stealthiness and attack success.

\subsection{Mechanism Verification}

To validate our hypothesis that the efficacy of GaussTrap stems from the adaptive proliferation of Gaussian primitives, we conduct a mechanism verification experiment by varying the densification limit $M$. Using the Earth image as the adversarial payload on the Blender dataset, we analyze how the primitive budget influences both backdoor fidelity and scene integrity. The quantitative results and visual evidence are summarized in Figure \ref{fig:combined_ablation}(B) and Figure \ref{fig:densification}.

As illustrated in Figure \ref{fig:densification}, we observe a clear positive correlation between the densification limit and attack success rate. A higher $M$ facilitates more frequent Gaussian fission, resulting in a dense population of fine-grained primitives. These "excess" Gaussians provide the high-dimensional capacity needed to locally encode complex adversarial patterns—effectively "carving" the backdoor into the spatial distribution without disrupting the global geometry of the benign scene. This confirms that GaussTrap exploits the over-allocation of primitives in 3DGS; by hijacking the model’s growth mechanism, adversarial information is strategically embedded into newly proliferated Gaussian clusters to ensure the attack is both high-fidelity and structurally entangled.

\subsection{\textbf{Ablation Study}}

\subsubsection{Stabilization Angle}

We performed an ablation study on our GaussTrap with stabilization angles using the Blender dataset with the Earth image as the attack image. As can be clearly observed from Figure \ref{fig:abu_angle}, applying VES significantly reduced artifacts around the attack viewpoint while improving the quality of rendered images near the attack viewpoint.

Moreover, the data in Table \ref{tab:AC} demonstrates that the use of VES enhances image quality under both training and testing viewpoints, thereby improving the overall performance of the model in terms of robustness and visual consistency. This clearly indicates that stabilization helps maintain continuity between attack viewpoints and normal viewpoints, preventing excessive distortion of the 3DGS model caused by the insertion of malicious viewpoints, and effectively mitigating the negative impact on the concealment of attacks.

\begin{table}[h!]
    \setlength{\tabcolsep}{5pt}
    \normalsize
    \centering
    \caption{Comparison of metrics with and without VES. The best performance is highlighted in \textbf{bold}.}
    \label{tab:AC}
    \begin{threeparttable}
    \resizebox{\linewidth}{!}{
    \begin{tabular}{lcccccc}
        \toprule
         & \multicolumn{3}{c}{\textbf{Train Viewpoints}} & \multicolumn{3}{c}{\textbf{Test Viewpoints}} \\
         \cmidrule(lr){2-4}\cmidrule(lr){5-7}
         & PSNR$\uparrow$ & SSIM$\uparrow$ & LPIPS$\downarrow$ & PSNR$\uparrow$ & SSIM$\uparrow$ & LPIPS$\downarrow$ \\
        \midrule
        \textbf{w/o VES} & 30.76 & 0.9642 & 0.0487 & 26.05 & 0.9070 & 0.0899 \\
        \textbf{w/ VES} & \textbf{32.56} & \textbf{0.9679} & \textbf{0.0433} & \textbf{27.98} & \textbf{0.9263} & \textbf{0.0727} \\
        \bottomrule
    \end{tabular}
    }
    
    \end{threeparttable}
\end{table}

\subsubsection{Attack viewpoints}
In real-world scenarios, attackers may attempt to inject malicious images from different perspectives to enhance the stealthiness and success rate of the attack. To evaluate the effectiveness of attacks from various perspectives, we uniformly selected $5$ different viewpoints and injected corresponding attack images accordingly, ensuring a diverse coverage of spatial angles. The quantitative analysis results are summarized in Figure \ref{fig:combined_ablation}(C). 

From the experimental results, we observe that although the image quality exhibits slight fluctuations as the viewpoint changes, it remains within an acceptable range across different evaluation perspectives. This indicates that GaussTrap can consistently support attack injections from multiple viewpoints without causing severe visual degradation. Such viewpoint adaptability further enhances the stealthiness of the attack, as the injected content can be embedded in a manner that remains visually plausible under different observation angles. These results demonstrate the robustness and practical applicability of GaussTrap in multi-view 3D reconstruction scenarios.

\begin{table}[h!]
    \setlength{\tabcolsep}{5pt}
    \normalsize
    \centering
    \caption{Comparison of the original poisoned model and the model after 1000-iteration fine-tuning defense.}
    \label{tab:defence_metrics}
    \begin{threeparttable}
    \resizebox{\linewidth}{!}{
    \begin{tabular}{lcccccc}
        \toprule
         & \multicolumn{3}{c}{\textbf{Train Viewpoints}} & \multicolumn{3}{c}{\textbf{Test Viewpoints}} \\
         \cmidrule(lr){2-4}\cmidrule(lr){5-7}
         & PSNR$\uparrow$ & SSIM$\uparrow$ & LPIPS$\downarrow$ & PSNR$\uparrow$ & SSIM$\uparrow$ & LPIPS$\downarrow$ \\
        \midrule
        \textbf{Poisoned} & 32.56 & 0.9679 & 0.0433 & 27.98 & 0.9263 & 0.0727 \\
\textbf{Fine-tuned} & 27.83 &	0.9338 &	0.0766 &	25.51 &	0.9046 &	0.0953 
 \\

        \bottomrule
    \end{tabular}
    }
    
    \end{threeparttable}
\end{table}

\begin{figure}[h] 
    \centering 
    \begin{tabular}{c@{\hspace{5pt}}c@{\hspace{5pt}}c} 
        \includegraphics[width=0.15\textwidth]{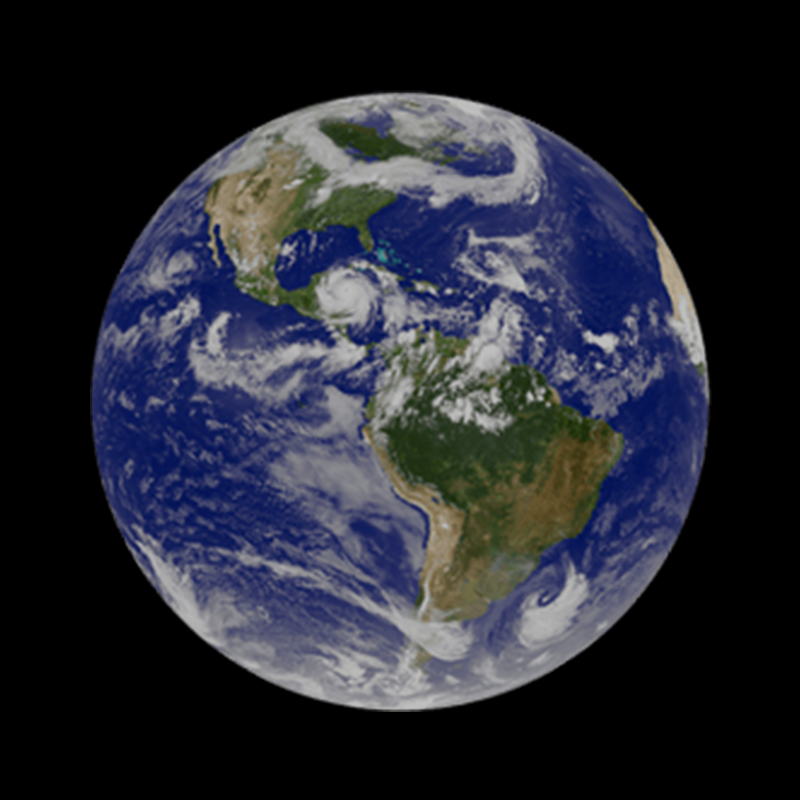} & 
        \includegraphics[width=0.15\textwidth]{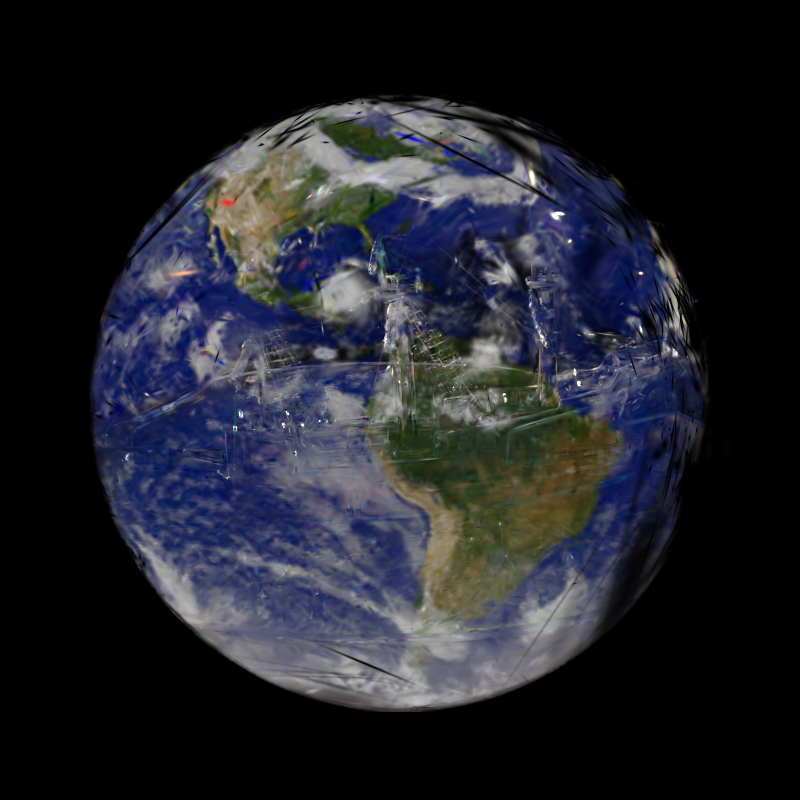} & 
        \includegraphics[width=0.15\textwidth]{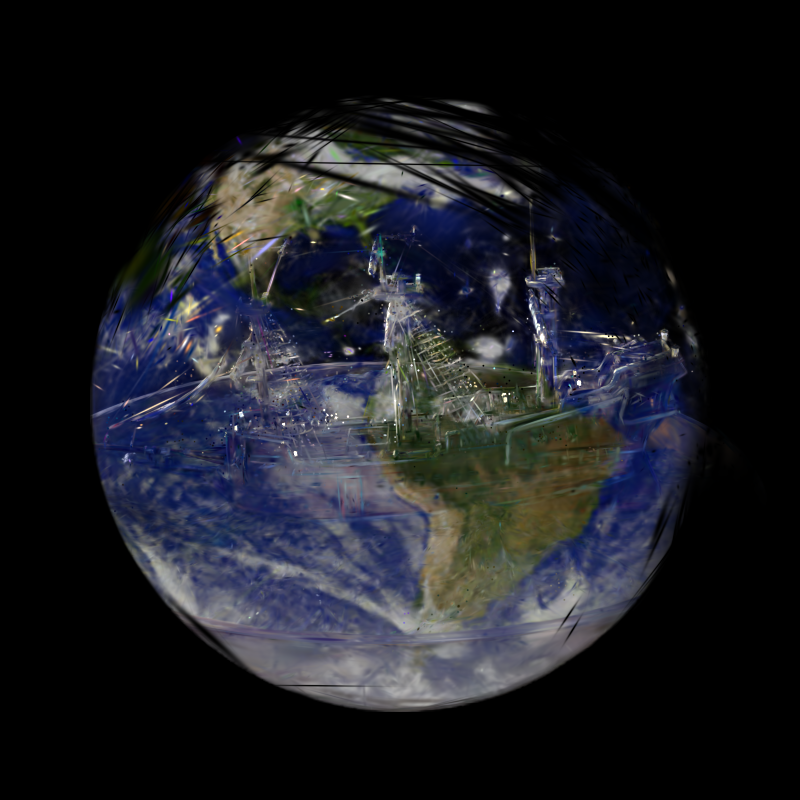} \\
        \small (a) Original & 
        \small (b) 500 Iterations&
        \small (c) 1000 Iterations\\
    \end{tabular}
        \caption{Visualization of images rendered from attack viewpoints during fine-tuning defense. } 
    \label{fig:defence_vis} 
\end{figure}

\subsection{Discussions on naive defense strategies}

A straightforward defense against GaussTrap is to fine-tune the backdoored 3DGS model on the original clean dataset, with the aim of overwriting the malicious content without explicitly identifying the poisoned viewpoints.

However, our experiments show that this simple fine-tuning strategy is ineffective. As shown in Figure~\ref{fig:defence_vis}, despite multiple rounds of fine-tuning, the model persistently renders the malicious content specified by the attacker from the designated attack viewpoint. This is because the injected backdoor is tightly entangled with the optimized 3D Gaussian parameters, making it difficult to remove through small-scale retraining.  Moreover, the image quality under normal viewpoints deteriorated during fine-tuning. The quantitative results in Table~\ref{tab:defence_metrics} further confirm this observation. The underlying reason is that fine-tuning disrupts the delicate balance established among the three stages of GaussTrap: adversarial injection, geometric stabilization, and global fidelity restoration. These findings underscore the need for more principled and effective defenses beyond naive fine-tuning.

\section{Conclusion}

In this work, we present \textbf{GaussTrap}, a novel and stealthy backdoor attack framework specifically designed for 3D Gaussian Splatting (3DGS). Our approach employs a systematic three-stage training pipeline—\textit{adversarial injection}, \textit{geometric stabilization}, and \textit{global fidelity restoration}—to embed malicious content that is deeply integrated into the model's structural parameters. By leveraging the intrinsic densification mechanism of 3DGS, GaussTrap achieves a superior balance between attack efficacy at trigger viewpoints and photorealistic fidelity in benign views. Extensive experiments on both synthetic and real-world datasets demonstrate that our method not only reliably induces targeted scene misperception but also exhibits significant resilience against naive defense strategies such as fine-tuning. These findings reveal a previously unexplored threat surface in neural scene representations, highlighting the urgent need for robust, structurally-aware defense mechanisms to safeguard emerging 3D rendering pipelines.

\begin{acks}
This work was supported in part by the National Natural Science Foundation of China under Grant Nos. 62301189, 62576122, and 62571298, and by the Guangdong Basic and Applied Basic Research Foundation under Grant No. 2026A1515011139.
\end{acks}

\bibliographystyle{ACM-Reference-Format}
\bibliography{kdd}

\clearpage
\appendix

\begin{figure*}[ht]
       \centering
    \begin{tabular}{cccc}
        \includegraphics[width=0.2\textwidth]{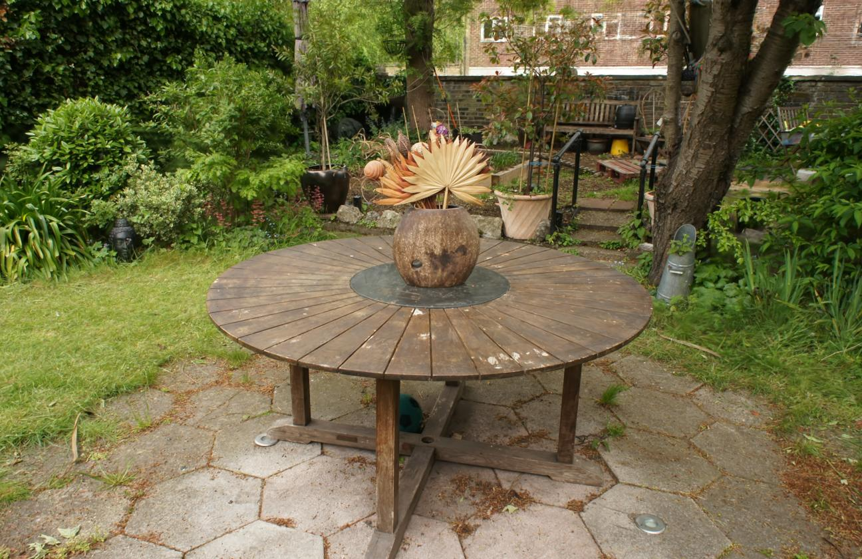} &
        \includegraphics[width=0.2\textwidth]{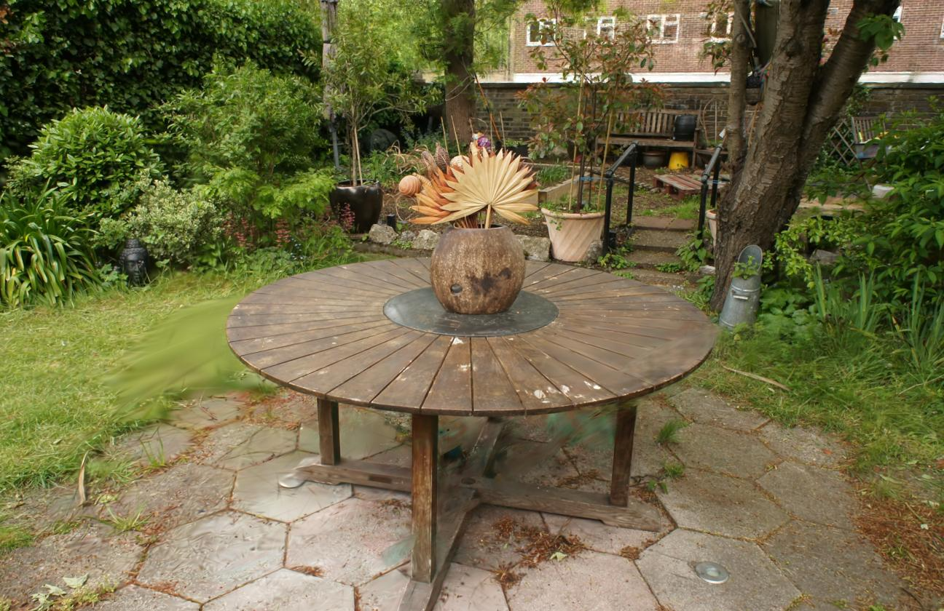} &
        \includegraphics[width=0.2\textwidth]{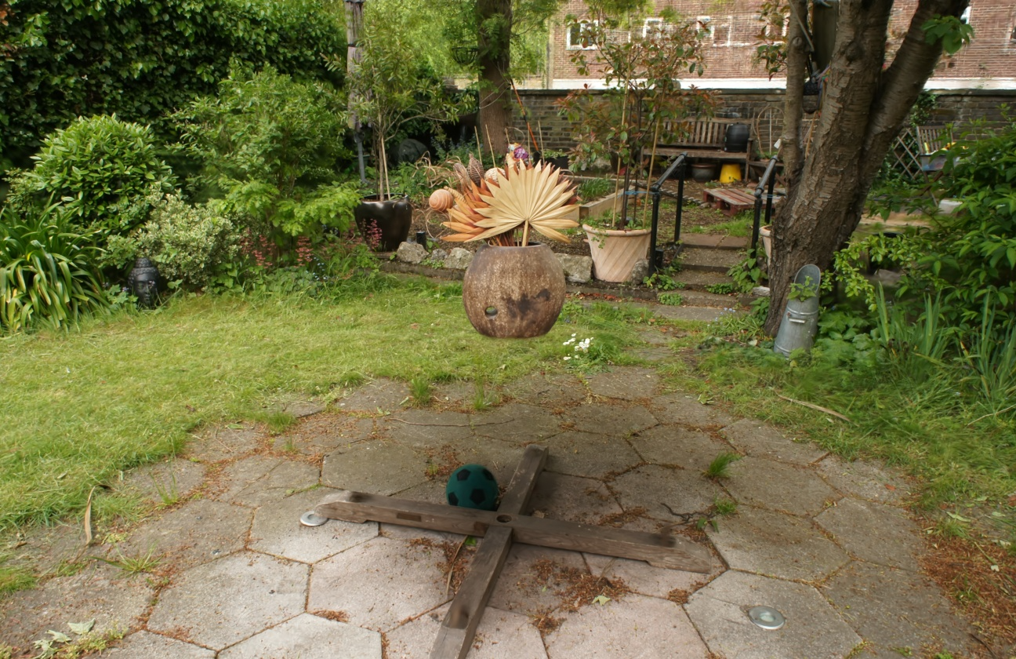} &
        \includegraphics[width=0.2\textwidth]{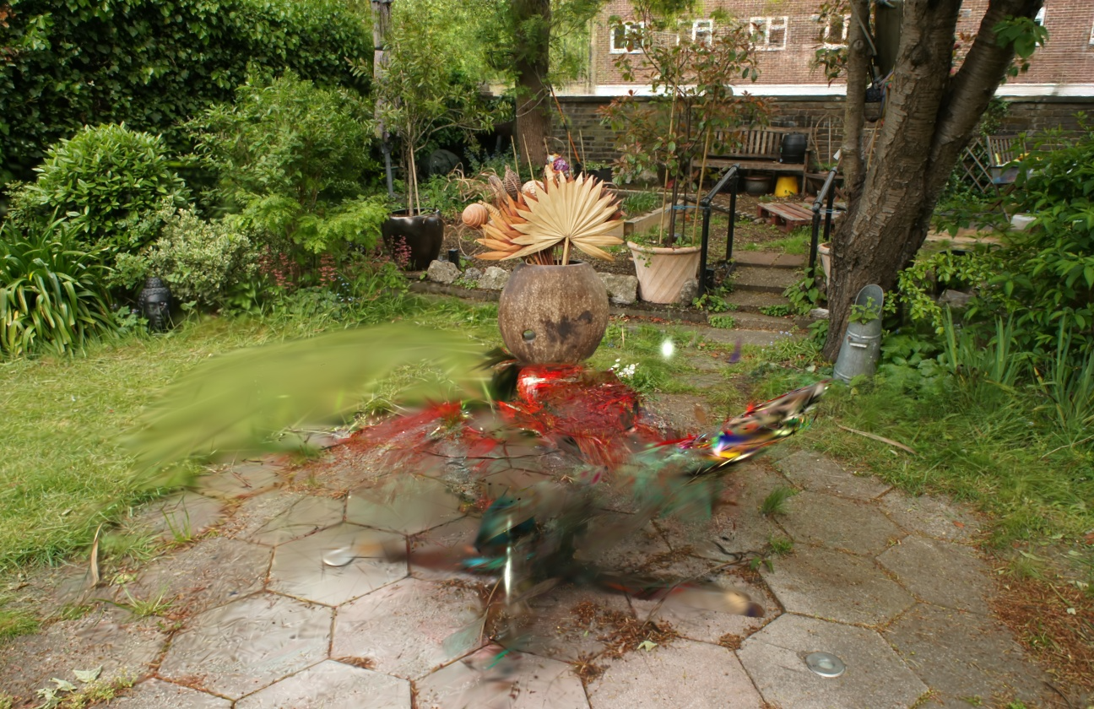} \\
        \includegraphics[width=0.2\textwidth]{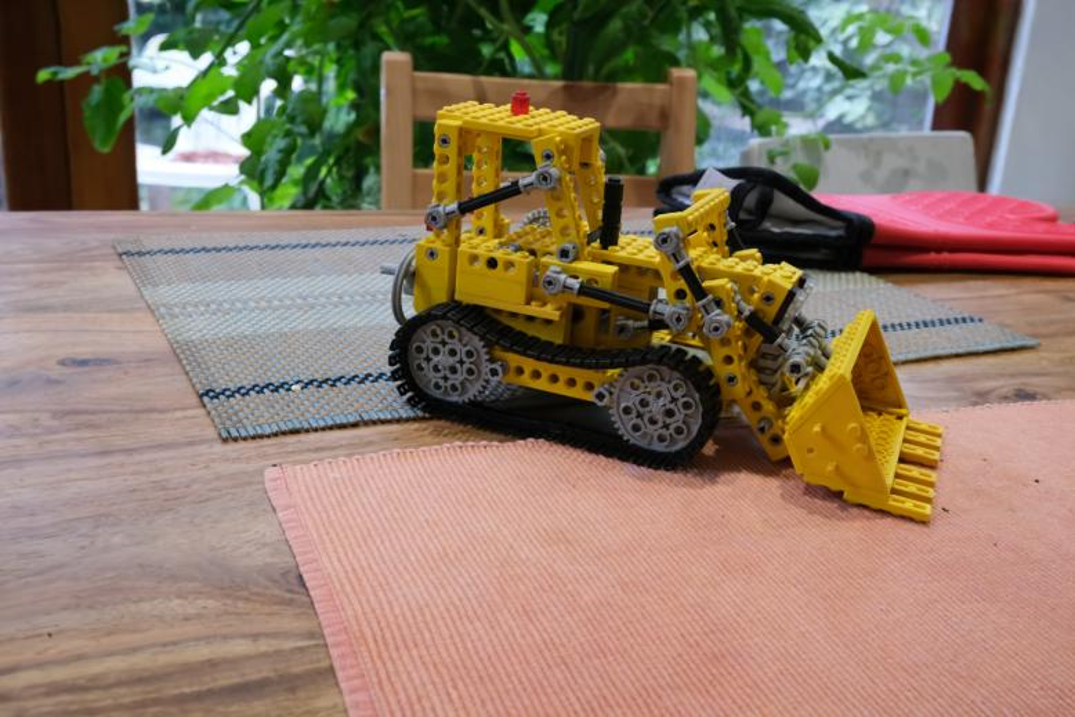} &
        \includegraphics[width=0.2\textwidth]{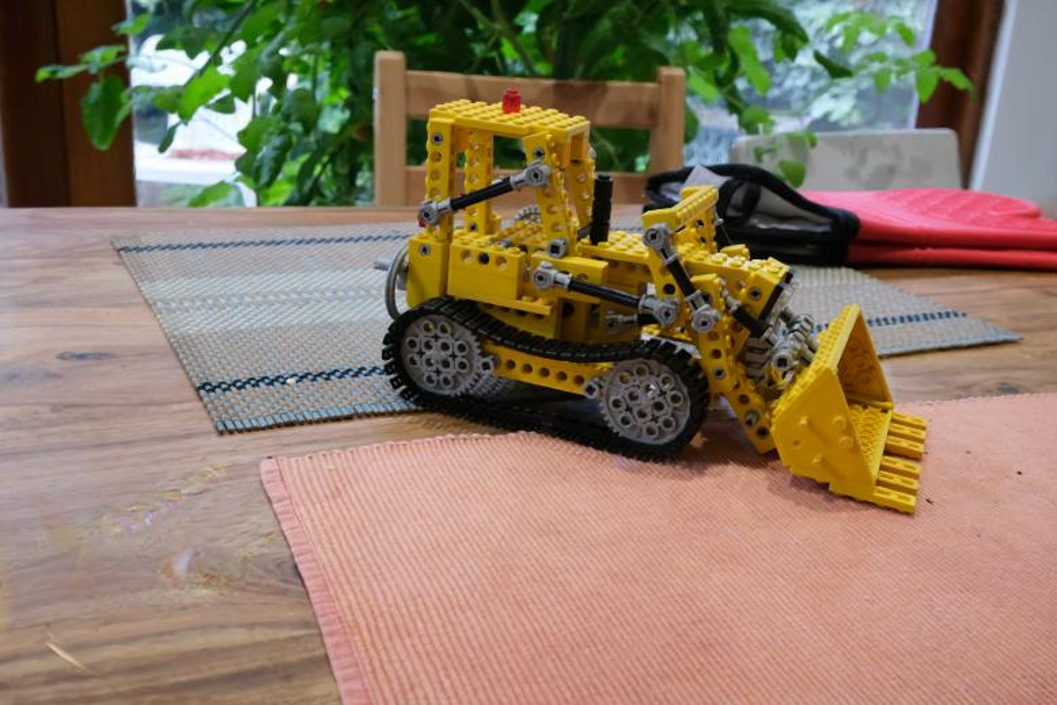} &
        \includegraphics[width=0.2\textwidth]{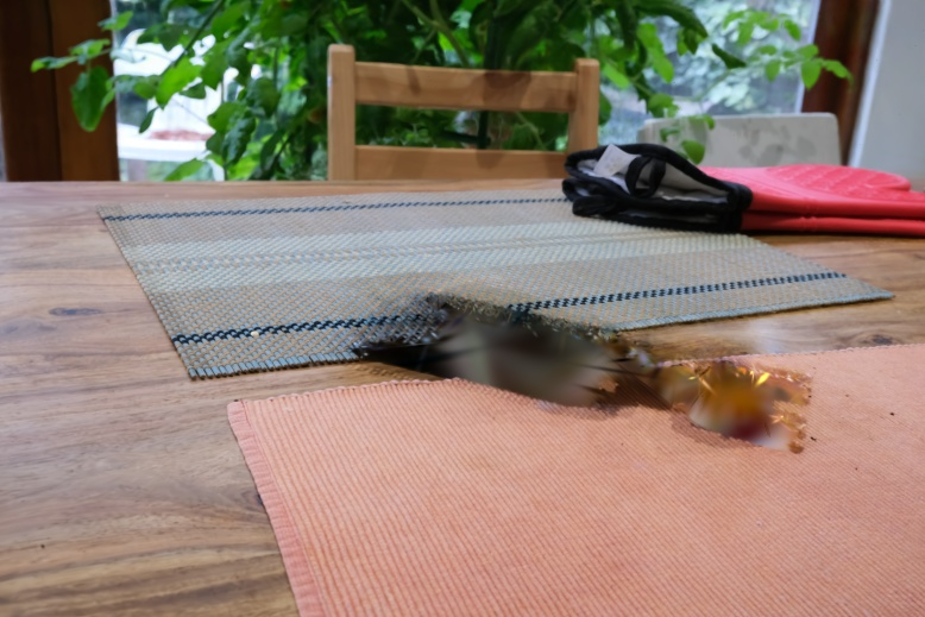} &
        \includegraphics[width=0.2\textwidth]{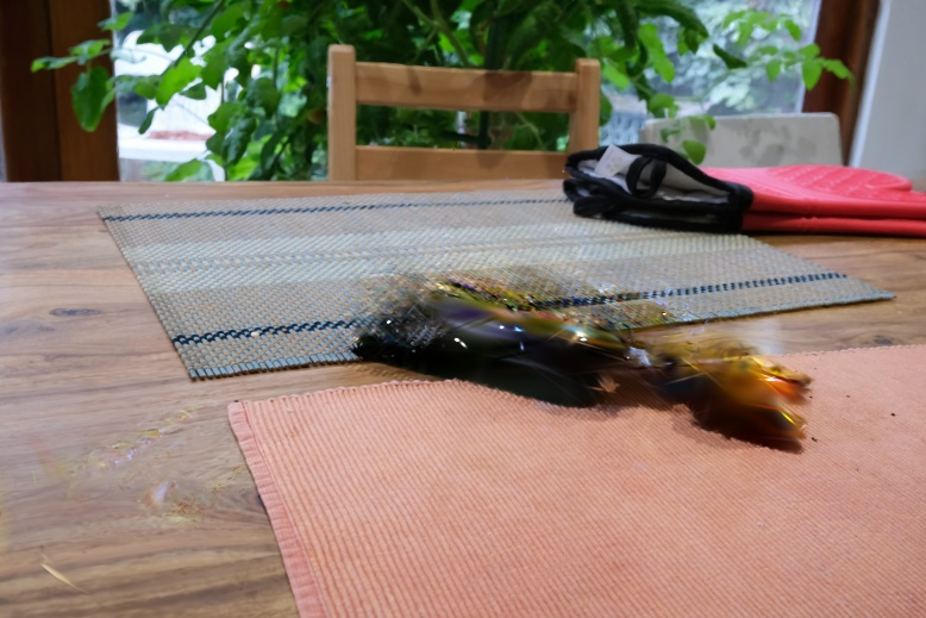} \\
            {\textbf{Clean 3DGS}} & 
            {\textbf{Poisoned 3DGS}} &
    {\textbf{Clean 3DGS + Removal}} & 
    {\textbf{Poisoned 3DGS + Removal}}  
    \end{tabular}
    \caption{Visualization of impact of GaussTrap on object removal on Mip-NeRF360 Dataset.}
    \label{fig:downstream}
\end{figure*}

\section{Additional Experimental Results}
\label{app:add_exp}


\subsection{Impact on Downstream Tasks}

We further evaluate GaussTrap in downstream 3DGS editing using a Gaussian scene segmentation and object removal framework~\cite{joji2024gradient}. As shown in Figure~\ref{fig:downstream}, the poisoned model appears normal before editing, but object removal introduces severe artifacts in normal views. This suggests that the backdoor is entangled with normal Gaussian components, causing downstream operations to corrupt previously clean renderings. Thus, GaussTrap threatens not only reconstruction integrity but also practical 3DGS applications such as editing, segmentation, and object removal.

\subsection{Defense and Detectability Analysis}
\label{app:defense}

\paragraph{Pruning-based defense.}
We evaluate whether the injected malicious Gaussians can be removed by simple pruning strategies. Specifically, we remove 40\% of the primitives using two pruning scores: global importance and view visibility. As shown in Table~\ref{tab:pruning_defense}, the attack remains effective after aggressive pruning, suggesting that the malicious Gaussians are entangled with normal scene geometry rather than being isolated outliers.

\begin{table}[ht]
\centering
\caption{Attack performance under pruning-based defenses with 40\% primitive removal.}
\label{tab:pruning_defense}
\begin{tabular}{lc}
\toprule
Pruning Strategy & Attack PSNR \\
\midrule
Global-importance & 30.52 \\
View-visibility & 30.46 \\
\bottomrule
\end{tabular}
\end{table}

We further evaluate LightGaussian-style significance-based pruning under the same 40\% pruning ratio. Table~\ref{tab:lightgaussian_pruning} shows that GaussTrap maintains stable attack performance under both global importance and view-visibility scores.

\begin{table}[ht]
\centering
\caption{Attack performance under LightGaussian-style significance-based pruning at 40\% pruning ratio.}
\label{tab:lightgaussian_pruning}
\begin{tabular}{lccc}
\toprule
Score Type & PSNR & SSIM & LPIPS \\
\midrule
Global Importance & 30.52 & 0.978 & 0.025 \\
View-Visibility & 30.46 & 0.977 & 0.025 \\
\bottomrule
\end{tabular}
\end{table}

\paragraph{Density-based anomaly filtering.}
We also test a simple density-based anomaly filter that removes the top 5\% densest Gaussian clusters. As shown in Table~\ref{tab:density_filtering}, the attack view still achieves high fidelity after filtering, while benign-view quality is largely preserved. This indicates that the malicious Gaussians cannot be easily identified as dense outlier clusters.

\begin{table}[ht]
\centering
\caption{Performance after removing the top 5\% densest Gaussian clusters.}
\label{tab:density_filtering}
\begin{tabular}{lc}
\toprule
Viewpoint Type & PSNR \\
\midrule
Attack View & 35.91 \\
Train View & 34.67 \\
Test View & 28.89 \\
Constraint View & 19.53 \\
\bottomrule
\end{tabular}
\end{table}

\subsection{Benchmark Diversity and Stronger Baselines}
\label{app:benchmark}

To further evaluate generality, we additionally conduct experiments on the Tanks\&Temples dataset. As shown in Table~\ref{tab:tanks_temples}, our method achieves substantially higher attack-view fidelity than prior baselines while maintaining competitive benign-view quality.

\begin{table}[ht]
\centering
\caption{Additional results on Tanks\&Temples.}
\label{tab:tanks_temples}
\begin{tabular}{lcccc}
\toprule
Method & Attack & Train & Test & Constraint \\
\midrule
IPA-NeRF + 3DGS & 26.64 & 22.11 & 20.94 & 31.79 \\
StealthAttack & 24.46 & 24.64 & 22.98 & 9.64 \\
Ours & 39.81 & 24.75 & 22.87 & 38.21 \\
\bottomrule
\end{tabular}
\end{table}

We also evaluate a direct transfer of BadNets-style 2D poisoning to 3DGS. As shown in Table~\ref{tab:badnets}, the attack PSNR is extremely low, suggesting that conventional 2D poisoning does not naturally transfer to 3DGS. This is likely because 2D triggers are disrupted by multi-view optimization and densification.

\begin{table}[ht]
\centering
\caption{Direct transfer of BadNets-style 2D poisoning to 3DGS.}
\label{tab:badnets}
\begin{tabular}{lc}
\toprule
Method & Attack PSNR \\
\midrule
BadNets + 3DGS & 4.73 \\
\bottomrule
\end{tabular}
\end{table}

\subsection{Hyperparameter Sensitivity}
\label{app:hyperparameter}

We study the sensitivity of GaussTrap to the iteration allocation among the Injection, Stabilization, and Restoration stages. Table~\ref{tab:iteration_ratio} reports the results under different iteration ratios. The results show a clear trade-off: stronger injection improves attack fidelity, while stronger restoration improves benign-view quality.

\begin{table}[ht]
\centering
\caption{Ablation on the iteration ratio among Injection, Stabilization, and Restoration stages.}
\label{tab:iteration_ratio}
\begin{tabular}{lcccc}
\toprule
I:S:R & Attack & Train & Test & Constraint \\
\midrule
35:5:15 & 40.48 & 29.50 & 23.78 & 26.40 \\
20:15:20 & 32.15 & 33.62 & 27.10 & 37.41 \\
25:5:25 (Ours) & 36.23 & 34.59 & 27.80 & 31.15 \\
\bottomrule
\end{tabular}
\end{table}

We observe that the attack can weaken when the trigger view is heavily occluded or far outside the training trajectory, where stabilization anchors are insufficient to constrain the local geometry.

\subsection{Viewpoint Robustness}
\label{app:viewpoint_robustness}

We conduct a finer-grained viewpoint perturbation analysis on Mip-NeRF360. Specifically, we perturb the camera around the trigger pose and use benign 3DGS renderings at the same perturbed poses as references. As shown in Table~\ref{tab:viewpoint_sensitivity}, the rendering gradually returns toward the benign scene as the viewpoint moves away from the trigger pose, rather than exhibiting abrupt transitions.

\begin{table}[ht]
\centering
\caption{Viewpoint perturbation analysis around the trigger pose on Mip-NeRF360.}
\label{tab:viewpoint_sensitivity}
\begin{tabular}{lccc}
\toprule
Deviation Angle & SSIM & PSNR & LPIPS \\
\midrule
1.0$^\circ$ & 0.7956 & 17.81 & 0.2282 \\
2.0$^\circ$ & 0.8096 & 19.15 & 0.2222 \\
3.0$^\circ$ & 0.8297 & 20.73 & 0.2083 \\
4.0$^\circ$ & 0.8529 & 22.36 & 0.1895 \\
\bottomrule
\end{tabular}
\end{table}

\subsection{Ablation of Global Fidelity Restoration}
\label{app:gfr_ablation}

We ablate the Global Fidelity Restoration stage to evaluate its contribution. As shown in Table~\ref{tab:gfr_ablation}, removing this stage substantially degrades rendering quality across all viewpoint types. This confirms that Global Fidelity Restoration is important for preserving benign scene fidelity while maintaining attack stealth.

\begin{table}[ht]
\centering
\caption{Ablation of Global Fidelity Restoration.}
\label{tab:gfr_ablation}
\begin{tabular}{lccc}
\toprule
Viewpoint Type & SSIM & PSNR & LPIPS \\
\midrule
Attack & 0.2527 & 9.67 & 0.4761 \\
Train & 0.2035 & 9.74 & 0.6264 \\
Test & 0.1975 & 10.10 & 0.6290 \\
Constraint & 0.2220 & 9.56 & 0.5670 \\
\bottomrule
\end{tabular}
\end{table}

\subsection{Model Size and Primitive Count}
\label{app:model_size}

We examine whether GaussTrap causes unnatural model bloat. As shown in Table~\ref{tab:model_size}, our method reduces both file size and point count compared with the original clean model. Although local densification introduces additional Gaussians near the attack region, this is offset by the global restoration stage and native opacity culling, which remove redundant background primitives.

\begin{table}[ht]
\centering
\caption{Comparison of model file size and point number.}
\label{tab:model_size}
\begin{tabular}{lcc}
\toprule
Model & File Size (MB) & Point Number \\
\midrule
Original & 71.82 & 972,896 \\
Ours & 66.38 & 726,768 \\
\bottomrule
\end{tabular}
\end{table}

\subsection{Training Time Overhead}
\label{app:training_time}

We report the training time overhead in Table~\ref{tab:training_time}. Our method introduces a moderate overhead compared with clean training, but is substantially more efficient than prior attack baselines. Specifically, GaussTrap is about $5\times$ faster than IPA-NeRF and about $7.5\times$ faster than StealthAttack.

\begin{table}[ht]
\centering
\caption{Training time comparison.}
\label{tab:training_time}
\begin{tabular}{lcc}
\toprule
Method & Time (Sec) & Overhead \\
\midrule
Original & 1,012 & 1.0$\times$ \\
Ours & 4,636 & 4.5$\times$ \\
IPA-NeRF & 23,264 & 23.0$\times$ \\
StealthAttack & 34,657 & 34.2$\times$ \\
\bottomrule
\end{tabular}
\end{table}

\subsection{Statistical Stability}
\label{app:std}

We further report standard deviations over 10 runs on Blender with the target image ``Earth''. As shown in Table~\ref{tab:std}, the standard deviations are consistently low across all viewpoint types and metrics, indicating stable convergence.

\begin{table}[ht]
\centering
\caption{Standard deviations over 10 runs on Blender.}
\label{tab:std}
\begin{tabular}{lccc}
\toprule
Viewpoint Type & PSNR SD & SSIM SD & LPIPS SD \\
\midrule
Attack & 0.12 & 0.001 & 0.001 \\
Train & 0.18 & 0.002 & 0.002 \\
Test & 0.21 & 0.003 & 0.003 \\
Stabilization & 0.15 & 0.002 & 0.002 \\
\bottomrule
\end{tabular}
\end{table}

\end{document}